# A non-linear learning & classification algorithm that achieves full training accuracy with stellar classification accuracy

Rashid Khogali
Department of Electrical & Computer Engineering
Ryerson University
Toronto, Canada
rkhogali@ryerson.ca, khogali@alumni.utoronto.ca

*Abstract-* A fast Non-linear and non-iterative learning and classification algorithm is synthesized and validated. This algorithm named the "Reverse Ripple Effect(R.R.E)", achieves 100% learning accuracy but is computationally expensive upon classification. The R.R.E is a (deterministic) algorithm that super imposes Gaussian weighted functions on training points. In this work, the R.R.E algorithm is compared against known learning and classification techniques/algorithms such as: the Perceptron Criterion algorithm, Linear Support Vector machines, the Linear Fisher Discriminant and a simple Neural Network. The classification accuracy of the R.R.E algorithm is evaluated using simulations conducted in MATLAB. The R.R.E algorithm's behaviour is analyzed under linearly and non-linearly separable data sets. For the comparison with the Neural Network, the classical XOR problem is considered.

**Contents**                                                                                                                                          **Pg.**

# 1 Introduction

In this study, a non-iterative algorithm named the "Reverse Ripple algorithm (R.R.E)" is synthesized and validated using numerical simulations conducted in MATLAB. The R.R.E algorithm is a classification algorithm that achieves 100% learning/training accuracy and stellar classification accuracy even with limited training data. This algorithm achieves stellar results when data is categorically separable (linearly as well as non-linearly separable).
The algorithm is modifiable such that it is able to:
- classify based on cost
- classify based on the multi-category case
- operate on an un-supervised mode after limited supervised training

These modifications above are not explored much due to the constrained scope of the paper. A major drawback of the algorithm is that it is computationally expensive upon classifying data when the training data is large hence it requires substantial further optimization.

## 1.1 Algorithm Validation

The R.R.E algorithm is compared to the P.C.A (Perceptron Criterion Algorithm [7]) using two types of datasets. It is then compared to Linear Support Vector Machines [2] on two types of data sets as well. Lastly, it is compared to a Neural Network ([3],[6]) of comparable complexity for a non-linearly separable case (XOR problem). Due to a time constraint imposed on the paper, Bootstrapping, Bagging, and Boosting validation techniques were not used, but instead, a variation of the S-fold cross validation technique was employed to validate the algorithm in comparison to the P.C.A. See [1]and [3] for details on the Perceptron Criterion Algorithm, Linear Support Vector machines, the Linear Fisher Discriminant ([5]) and Neural Networks.

# 2 Introducing the (R.R.E) Algorithm

## 2.1 Brainstorming

Let us attempt to superimpose any monotonically decreasing weight function at all training set points, intuitively this makes sense as it increases the weight of properly classifying a test point that is radially close to a known training point. The idea is we are using training points as our evidence and associating a modified distance (dissimilarity) between a test point and neighbourhood training points as our criteria of categorizing a test point. The smaller the distance or dissimilarity, it is more likely that a test point belongs to a region that is dominated or heavily populated by the training data in question and the larger the distance, the unlikely the training point belongs to that region; this explains why the weight functions have to monotonically decrease as directed by our intuition.

We also know that initially our training set is small, so the variances of our weight functions should be huge to maximize the known entropy/known information even if it means that we may poorly classify regions that are very far from training data - this is the price we pay for having little information. As our training set grows, we can afford to reduce the variance of our weight functions and make more accurate decisions based on smaller distances that are associated with high confidence levels. Let us introduce mathematical syntax to better capture the idea being proposed.

## 2.2 Preliminary Definitions

Due to the limited scope of the paper, we only consider the two category case, but the reader is advised that the algorithm can effortlessly be extended to the multi-category case.

Let $T_1$ and $T_2$ be the set of all training points associated with category $W_1$ and category $W_2$ respectively, where $x_i \in T_1$ and $x_j \in T_2$. Let $V_k$ be the set of all test/verification data where $x_k \in V_k$.

## 2.3 The Monotonically Decreasing Weight Function

Let $W_m(x)$ be the monotonically decreasing weight function. A good hunch is to let $W_m(x)$ be Gaussian, but other desirable choices would be the decay exponential or the hyperbolic function. We stick to the Gaussian for now because the central limit theorem tells us that the sum of I.I.D random variables converge to a Gaussian; if we treat a specific training point as a Bernoulli R.V with respect to categorization ($w_1$ or $w_2$), without any prior information, it can have us to believe that the sum of many future testing points that converge in the neighbourhood or vicinity of that training point may behave more and more like the normal distribution.

So $W_m(x) = e^{-x^2}$.

Next we superimpose $W_m(x)$ at all training set points. From signal processing, this operation is a convolution operation between $W_m(x)$ and a delta functions centred at testing points. Superposition is a great idea because it is a suitable way to consider multiple responses simultaneously.
So the category 1 is classified by the influence of:
$$W_m(x) \star \sum_i \delta(x - x_i), \quad \forall x_i \in T_1 \qquad (1)$$

Likewise, the category 2 is classified by the influence of:
$$W_m(x) \star \sum_j \delta(x - x_j) \quad \forall x_j \in T_2 \qquad (2)$$

(1) can be written as: $\sum_i W_m(x - x_i), \quad \forall x_i \in T_1$ \qquad (3)

(2) can be written as: $\sum_j W_m(x - x_j), \quad \forall x_j \in T_2$ \qquad (4)

We can write the equations above in more explicit forms.
Writing (3) in more explicit form is: $\sum_i e^{-(x-x_i)^t(x-x_i)}, \quad \forall x_i \in T_1$

Writing (4) in more explicit form is: $\sum_j e^{-(x-x_j)^t(x-x_j)}, \quad \forall x_j \in T_2$

We can now come up with a skeleton Discriminant function G(x) that takes the testing points $x_k$ as its argument.

$$\boxed{G(x_k) = \sum_i e^{-(x_k-x_i)^T(x_k-x_i)} - \sum_j e^{-(x_k-x_j)^T(x_k-x_j)}, \quad \forall x_i \in T_1, \forall x_j \in T_2 \text{ and } x_k \in V_k}$$

- If $G(x_k) > 0$ we categorize $x_k$ as category $w_1$
- If $G(x_k) < 0$ we categorize $x_k$ as category $w_2$
- If $G(x_k) = 0$, we reject $x_k$ or we toss a coin (a joke)

## 2.4 Introducing Cost (P)
The nice thing about our Discriminant function above is it allows us to introduce cost in so many ways. Let us keep it simple. Assuming a linear fixed cost, **p₁** associated with choosing **w₁** and a linear fixed cost **p₂** associated with choosing category **w₂**, we can modify our Discriminant function as follows:

$$G(x_k) = p_2 \sum_i e^{-(x_k-x_i)^T(x_k-x_i)} - p_1 \sum_j e^{-(x_k-x_j)^T(x_k-x_j)}, \quad \forall x_i \in T_1, \forall x_j \in T_2 \text{ and } x_k \in V_k$$

Notice that $p_2$ enters the influencing factor for category 1 and vice versa because a high linear cost of category 2 should boost the picking/selection of category 1.

Depending on the nature of the cost, it can also be simultaneously or separately introduced in the exponential argument, but we will ignore this for now, but keep in mind that we can introduce multiple types of costs simultaneously with different degrees of influence.

## 2.5 Introducing Variance

Let $n_i$ be the number of $\forall x_i \in T_1$ and let $n_j$ be the number of $\forall x_j \in T_2$.

The reason we split $n_i$ and $n_j$ instead of using $n_i + n_j$ is because we need to give a fair chance to either category in the event that the training data is biased (we are indirectly mitigating (potential) sample-model error).

Next, let f(n) be any increasing function of n. we are trying to grow f(n) with n. The exact family of functions for n that does a good job is unknown for now and suggest it could possibly be derived empirically. So we will leave f(n) in its implicit form.

Introducing a dynamic variance as opposed to assuming a fixed variance, we modify the Discriminant function as follows:

$$G(x_k) = p_2 \sum_i e^{-f(n_i)(x_k-x_i)^T(x_k-x_i)} - p_1 \sum_j e^{-f(n_j)(x_k-x_j)^T(x_k-x_j)}, \quad \forall x_i \in T_1, \forall x_j \in T_2 \text{ and } x_k \in V_k$$

## 2.6 Introducing the Auxiliary Sensitivity Factor (λ)

When we are not concerned about over-fitting, but we want our Discriminant function to obtain outputs that are arbitrarily close to our training data, we can use a sensitivity $\lambda \gg 1$. In the modified Discriminant function below:

$$G(x_k) = p_2 \sum_i e^{-\lambda f(n_i)(x_k-x_i)^T(x_k-x_i)} - p_1 \sum_j e^{-\lambda f(n_j)(x_k-x_j)^T(x_k-x_j)}, \quad \forall x_i \in T_1, \forall x_j \in T_2 \text{ and } x_k \in V_k$$

The larger the $\lambda$, the less we take advantage of mutual information between neighbouring data points and each training point becomes more isolated like a delta function, hence we are able to retrieve any training value with minimal interference from neighbouring training values. For much of this paper, we will eliminate the sensitivity factor and assume a unitary value for now to avoid over fitting. We will reintroduce it again in section 3.5 to demonstrate that the algorithm is capable of achieving 100% learning/training accuracy with an adjusted $\lambda$.

## 2.7 Prediction Capabilities of Discriminant Function

The Discriminant function of R.R.E algorithm has a quantitative interpretation. Under unitary categorical cost, when evaluated at a test point, it tells us roughly the number of training points(and/or fraction of a training point)that lie in the 'vicinity' of a test point. The larger the absolute value of the Discriminant function evaluated at a test point, the better the confidence level of the prediction/classification. In fact, in-light of this insight, we can choose to discard output values of the Discriminant functions that do not meet a reliable threshold, where the threshold can be empirically derived depending on the application. Implementing this threshold would improve the accuracy of classification but introduces a rejection set – which warrants require further consideration.

## 2.8 Naming & Defining the Classification Algorithm

Lets name the algorithm the 'Reverse-ripple Effect' algorithm (R.R.E) because each training point is like a 'water ripple' with a Gaussian like function superimposed on it, but unlike a water ripple, its variance becomes smaller as more training points are introduced. i.e the radial Gaussian variance decreases hence 'reverses' unlike a true ripple seen in water which actually expands with time.

The R.R.E algorithm is a multi-part algorithm. For most classification applications, we only require Part(A) followed by Part(C).

(A) Training:
- Start of with some training sets $T_1$ and $T_2$.
- Prepare $x_k$ (test point)
- Find $n_1$ and $n_2$ (number of points in $T_1$ and $T_2$ respectively.
- Evaluate $f(n_1)$ and $f(n_2)$ (can assume $f(n) = n$ and $\lambda=1$)
- Construct the Discriminant below:

$$G(x_k) = p_2 \sum_i e^{-\lambda f(n_i)(x_k-x_i)^T(x_k-x_i)} - p_1 \sum_j e^{-\lambda f(n_j)(x_k-x_j)^T(x_k-x_j)}, \forall x_i \in T_1, \forall x_j \in T_2 \text{ and } x_k \in V_k$$

(B) Extra supervised training (rarely needed in practice)
- use $G(x_T)$ above to classify all $x_T \in T_1 \bigcup T_2$
- If any $x_T$ is misclassified, redundantly include $x_T$ in the correct training set (in appropriate T₁ or T₂) and/or adjust $\lambda$ (the auxiliary sensitivity factor)
- Re-construct $G(x_k)$ with updated T₁ and T₂
- Stop when $G(x_T)$ correctly classify all $x_T \in T_1 \bigcup T_2$

(C) Classifying (predictor)
  Use $G(x_k)$ to classify any or all $x_k \in V_k$

(D) Classifying with further (unsupervised) learning
- Use $G(x_k)$ above to classify $x_k \in V_k$
- put x_k in T₁ if x_k was classified as w₁ and otherwise, put x_k in T₂.
- Re-construct $G(x_k)$ with updated T₁ and T₂
- Repeat for all $x_k \in V_k$

Note for Part D only: The memory requirements grow linearly with every test point that is included in the training set and can be stopped once sufficient test data has been incorporated into training data.

**For most classification applications and for the rest of this paper, we safely assume that the R.R.E Algorithm is Part(A) followed by Part(C) only.**

## 2.9 Potential Optimization
2.9.1: The possible need for 'filtering' the training data
The memory requirements of the Discriminant function grows linearly with every training point that it uses. Classification becomes computationally expensive when the training set $T_1 \bigcup T_2$ is large. One way to fix this without much elaboration is; after part A or Part B of the algorithm, for any training point, temporarily remove it from the Discriminant function, run the Discriminant function and see if it correctly classifies **all original** training points including the training point that was removed. If it does, then the removed training point was redundant, so we can safely eliminate ('filter') it permanently from our Discriminant function. We have to iteratively do this for all our training points and our final reduced training points that remain in the Discriminant function may not be unique because this filtering process is sensitive to the order in which we select the training points that we wish to filter. This filtering exercise has a computational complexity of $O(n^2)$, (where n is the number of training points)and may be worth while if the number of test data is substantial.

## 2.10 R.R.E Algorithm Impervious to Noisy Training Environments

We define noisy training environments are such that there is a high cross correlation or better yet, similarity between training points of different categories. More precisely, a noisy training environment is: if $x_i \in T_1$ and $x_j \in T_2$, there exist some or many training points such that $x_i \cong x_j$. In a noisy training environment, as we expand our training set by adding training data points that we know are correct, and in the rare case that a new training point has a categorization that conflicts with the classification of the Discriminant function, we simply include the point m times. m for the worse case scenario is usually 2 indicating that it was initially used to cancel the wrong pre-existing training point and the other is to fully guarantee the correct classification training of the more recent training point in question. This is very powerful because noisy data can be cancelled out when m = 1 and recent, more correct training is enforced when m = 2. Note that m = 2 will rarely be needed as duplication is rare (we rarely have the existence of any $x_i = x_j$), so most of the time, m = 1 rectifies and enforces training accordingly. To mathematically describe what has been discussed above, Let us initially have a noisy training point $\hat{x}_f \in T_1$ that is initially included in our Discriminant function at some early stage in the training. (Assuming equal categorical cost: $p_1 = p_2$.)

$$G_{(1)}(x_k) = p_2 \sum_i e^{-f(n_i)(x_k - x_i)^T (x - x_i)} - p_1 \sum_j e^{-f(n_j)(x_k - x_j)^T (x_k - x_j)} \quad : \hat{x}_f, \forall x_i \in T_1, \forall x_j \in T_2 \text{ and } x_k \in V_k$$

$$= p_2 e^{-f(n_i)(x_k - \hat{x}_f)^T (x - \hat{x}_f)} + p_2 \sum_{i \neq f} e^{-f(n_i)(x_k - x_i)^T (x_k - x_i)} - p_1 \sum_j e^{-f(n_j)(x_k - x_j)^T (x_k - x_j)}$$

Later as we train, we discover that due to noisy training data, we have an ambiguity because $\hat{x}_f \in T_2$, we implement training as normal and we have

$$G_{(2)}(x_k) = G_{(1)}(x_k) - p_1 e^{-f(n_i)(x_k - \bar{x}_f)^T (x_k - \bar{x}_f)}$$

$$G_{(2)}(x_k) = p_2 e^{-f(n_i)(x_k - \bar{x}_f)^T (x_k - \bar{x}_f)} + p_2 \sum_{i \neq f} e^{-f(n_i)(x_k - x_i)^T (x - x_i)} - p_1 e^{-f(n_i)(x_k - \bar{x}_f)^T (x - \bar{x}_f)} - p_1 \sum_{j \neq f} e^{-f(n_j)(x_k - x_j)^T (x_k - x_j)}$$

$$= p_2 e^{-f(n_i)(x_k - \bar{x}_f)^T (x_k - \bar{x}_f)} - p_1 e^{-f(n_i)(x_k - \bar{x}_f)^T (x_k - \bar{x}_f)} + p_2 \sum_{i \neq f} e^{-f(n_i)(x_k - x_i)^T (x_k - x_i)} - p_1 \sum_{j \neq f} e^{-f(n_j)(x_k - x_j)^T (x_k - x_j)}$$

$$= p_2 \sum_{i \neq f} e^{-f(n_i)(x_k - x_i)^T (x_k - x_i)} - p_1 \sum_{j \neq f} e^{-f(n_j)(x_k - x_j)^T (x_k - x_j)}$$

Note $p_2 e^{-f(n_i)(x_k - \bar{x}_f)^T (x - \bar{x}_f)} - p_1 e^{-f(n_i)(x_k - \bar{x}_f)^T (x - \bar{x}_f)} = 0$ because we initially assumed equal categorical cost: $p_1 = p_2$, so algebraically, we have demonstrated the desirable cancellation of the noisy training data point $\hat{x}_f$.

If we now want to enforce the training of $\hat{x}_f \in T_2$
We have

$$G_{(3)}(x_k) = G_{(2)}(x_k) - p_1 e^{-f(n_i)(x_k - \bar{x}_f)^T (x_k - \bar{x}_f)}$$

$$G_{(3)}(x_k) = p_2 \sum_{i \neq f} e^{-f(n_i)(x_k - x_i)^T (x_k - x_i)} - p_1 \sum_{j \neq f} e^{-f(n_j)(x_k - x_j)^T (x_k - x_j)} - p_1 e^{-f(n_i)(x_k - \bar{x}_f)^T (x_k - \bar{x}_f)}$$

$$= p_2 \sum_{i \neq f} e^{-f(n_i)(x_k - x_i)^T (x_k - x_i)} - p_1 \sum_{j} e^{-f(n_j)(x_k - x_j)^T (x_k - x_j)} \quad \forall x_i \in T_1, \hat{x}_f, \forall x_j \in T_2 \text{ and } x_k \in V_k$$

Note: this convenient 'cancellation' of noisy training points may not work well when we have an unequal categorical cost; that is when $p_1 \neq p_2$. To be more precise, if initially $\hat{x}_f \in T_1$, but we later want to enforce $\hat{x}_f \in T_2$, training noise cancellation:
- fails if $p_1 < p_2$.
- works if $p_1 > p_2$ with an unnecessary over-compensation.
- works perfectly if $p_1 = p_2$.

### 2.11 R.R.E Algorithm Validation
The R.R.E algorithm is compared to the P.C.A (Perceptron Criterion Algorithm) in section 3 using two types of datasets. It is then compared to Linear Support Vector Machines in section 4 on two types of data sets as well. Lastly, in section 5, it is compared to a Neural Network of comparable complexity under a simple non-linearly separable case (XOR problem).

## 3 Comparing R.R.E to P.C.A

### 3.1: Using "iris setosa versicolor" Dataset

"iris_setosa_versicolor" data is an appropriate two-category dataset that exhibits the strength of P.C.A (Perceptron Criterion Algorithm) because the data is somewhat linearly correlated along each category. The data is used to demonstrate and contrast the performance of both the P.C.A and R.R.E algorithms.

The figure below visually depicts the "iris_setosa_versicolor" dataset. Refer to Appendix B1 to view a tabular listing of all normalized and augmented row vectors contained in the dataset. We can see the data is somewhat linearly correlated along each category.

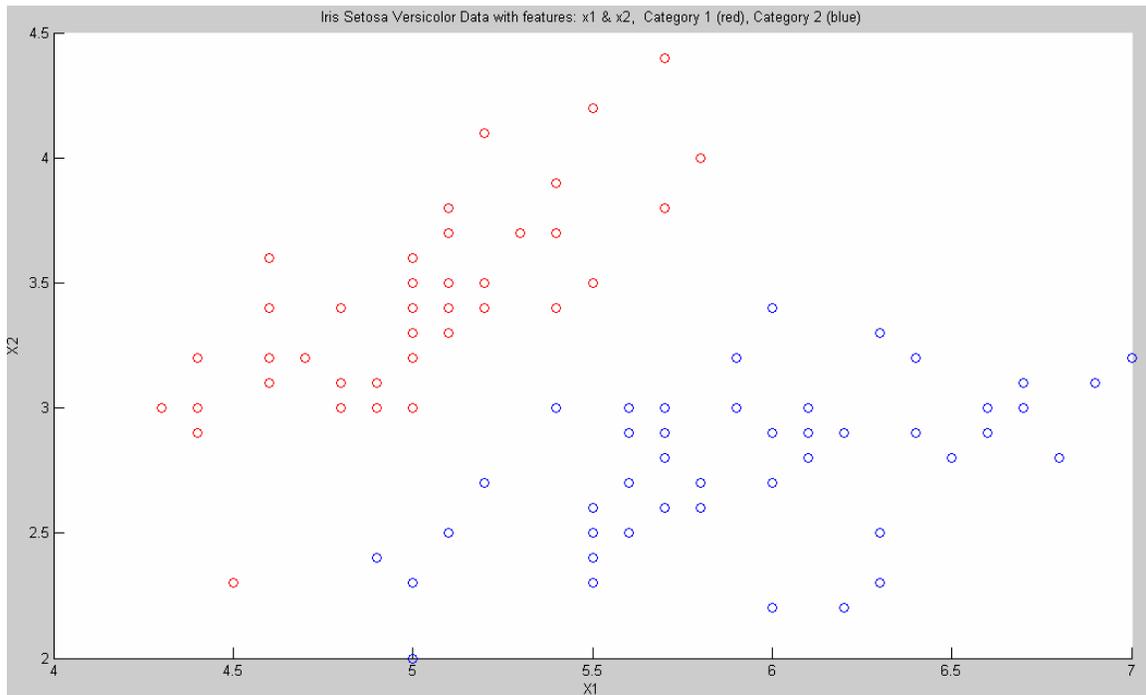

**Figure 3.1**: Visual representation of "iris_setosa_versicolor" dataset

### 3.2: Applying R.R.E to the "iris setosa versicolor" Dataset

**3.2(a):** 40% Training Data & 60% Test Data
The dataset of ("iris_setosa_versicolor") was split into two sets. We placed the first 40% of the data in one set (training set) and the remaining 60% (test set) in the second set.
We start of with the R.R.E algorithm Discriminant function:

$$G(x_k) = p_2 \sum_i e^{-f(n_i)(x_k-x_i)^T(x_k-x_i)} - p_1 \sum_j e^{-f(n_j)(x_k-x_j)^T(x_k-x_j)} \quad \forall x_i \in T_1, \ \forall x_j \in T_2 \ \text{and} \ x_k \in V_k$$

Assuming equal unitary categorical cost (p₁ = p₂) and a variance reduction function of $f(n_i) = f(n_j) = n_i = n_j = 20$, our Discriminant function reduces to:

$$G(x_k) = \sum_i e^{-20\,(x_k-x_i)^T(x_k-x_i)} - \sum_j e^{-20\,(x_k-x_j)^T(x_k-x_j)} \quad \forall x_i \in T_1,\ \forall x_j \in T_2 \text{ and } x_k \in V_k$$

After running the test data on the Discriminant function above, we get a classification summarized by the table below.

**TABLE 3.2a**: Classification summary of R.R.E algorithm applied to "iris_setosa_versicolor" using 40% training data and 60% test data.

| Category | Number of training vectors Misclassified | Number of testing vectors Misclassified | Classifier Accuracy excluding Training Data | Classifier Accuracy including Training Data |
|---|---|---|---|---|
| one | 0/20 | 1/30 | 96.67% | 98% |
| two | 0/20 | 0/30 | 100% | 100% |
| both | 0/40 | 1/60 | **98.33%** | 99% |

The decision surface of the configuration above (using 40% training data and 60% test data) is provided below.

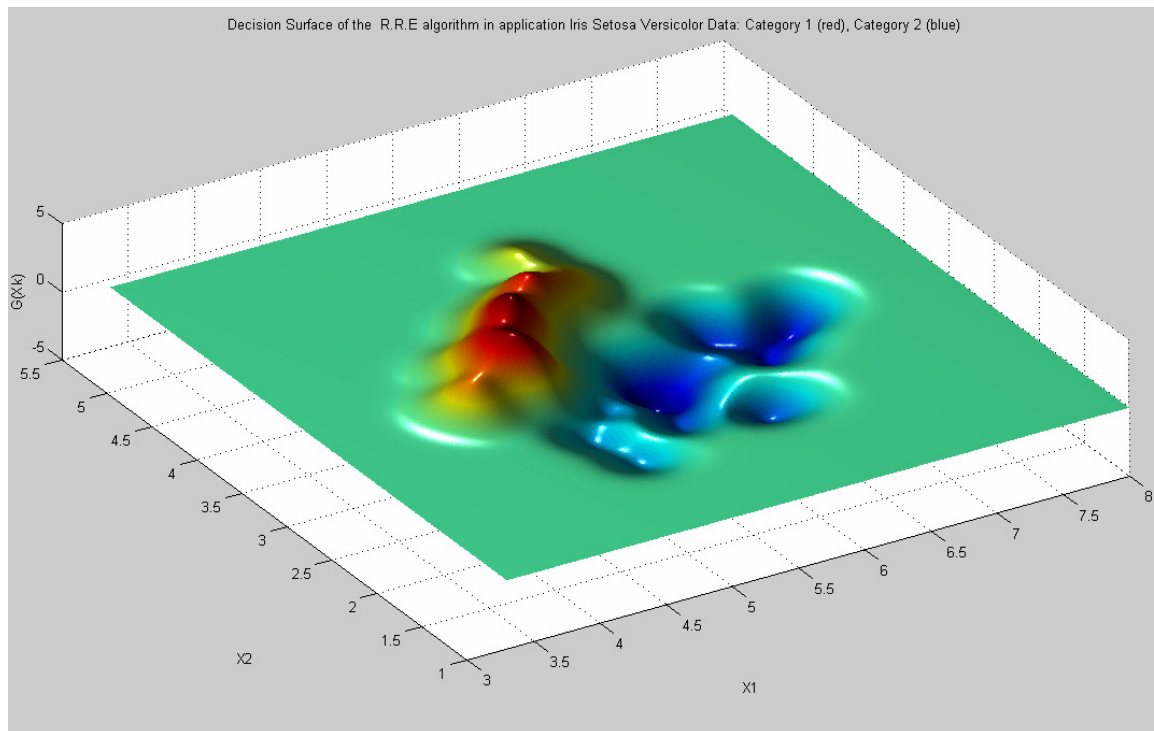

**Figure 3.2a**: Decision surface of R.R.E algorithm applied to "iris_setosa_versicolor" using 40% training data and 60% test data.

The decision surface shown above indeed shows the behaviour of the algorithm's Discriminant function, but what is of essential interest is the decision boundary because it visually depicts the separation of decision regions. In the contour diagram below, the decision boundary is seen when the contour height is zero (indicated by the green line labelled'0') which is equivalent to the Discriminant function being zero.

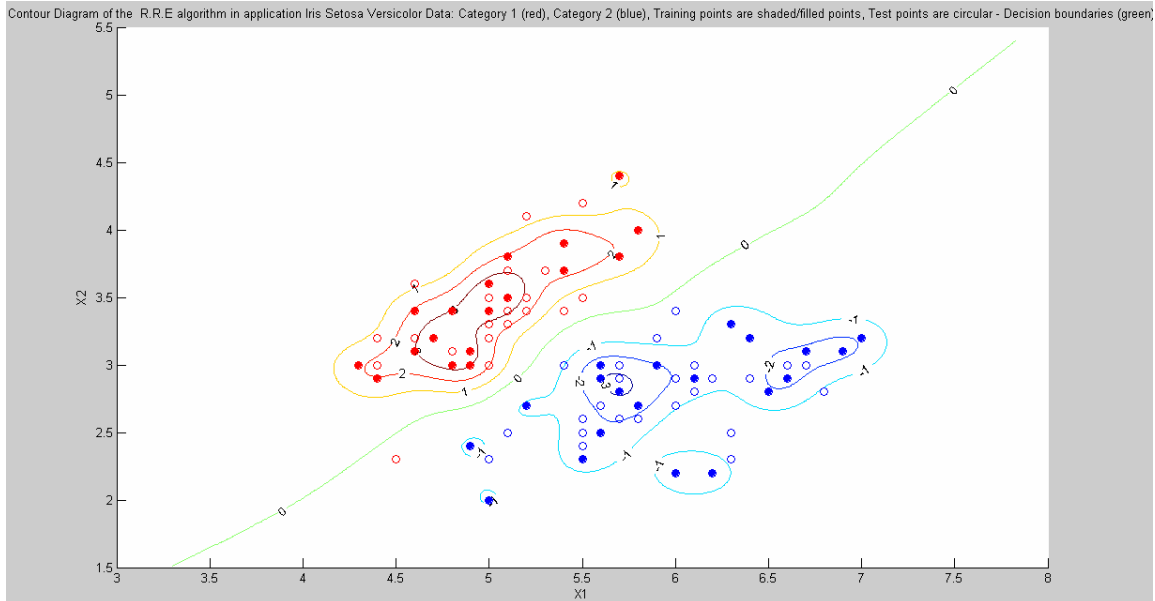

**Figure 3.2a2**: Contour Diagram of R.R.E algorithm applied to "iris_setosa_versicolor" using 40% training data and 60% test data.
Note: decision boundary indicated by '0' (zero contour height). We have a parasitic outlier at $(4.5, 2.3)^T$ that is solely misclassified.

### 3.2(b): 60% Training Data and 40% Test Data
The dataset of ("iris_setosa_versicolor") was split into two sets with the first 60% of the data in one set(training set) and the remaining 40%(test set) in the second set. Repeating section 3.2(a) with 60% training data and 40% test data, we obtain a classification summarized by the table below.

**TABLE 3.2.b**: Classification summary of R.R.E algorithm applied to "iris_setosa_versicolor" using 60% training data and 40% test data.

| Category | Number of training vectors Misclassified | Number of testing vectors Misclassified | Classifier Accuracy excluding Training Data | Classifier Accuracy including Training Data |
|----------|------------------------------------------|------------------------------------------|---------------------------------------------|---------------------------------------------|
| one      | 0/30                                     | 1/20                                     | 95.00%                                      | 98%                                         |
| two      | 0/30                                     | 0/20                                     | 100%                                        | 100%                                        |
| both     | 0/60                                     | 1/40                                     | **97.5%**                                   | 99%                                         |

Similarly, the decision surface of the configuration above (using 60% Training data and 40% test data) is presented below.

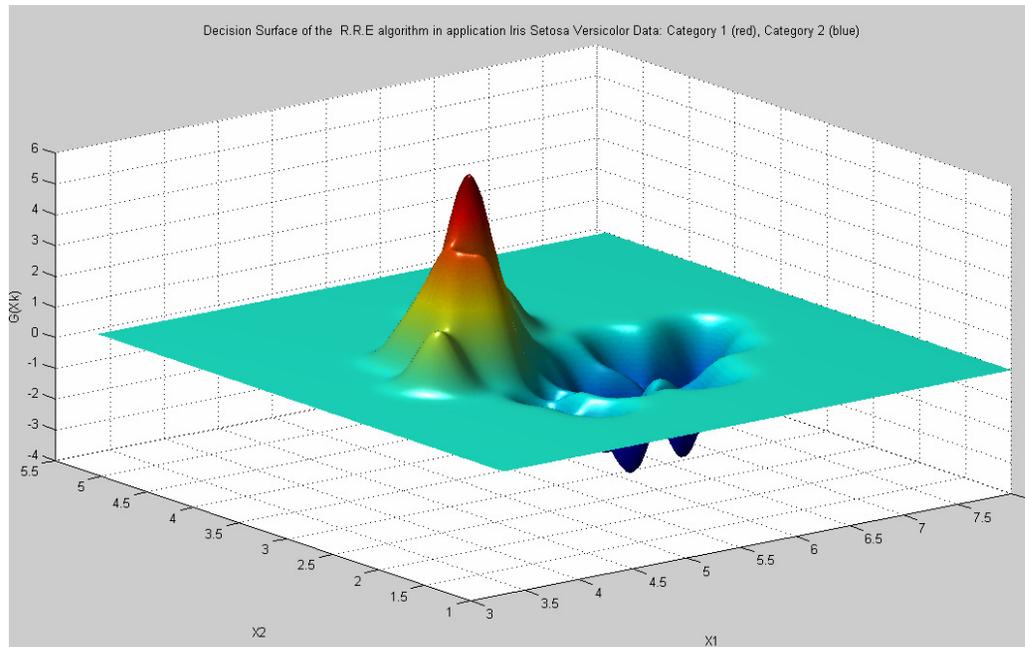

**Figure 3.2.b**: Decision surface of R.R.E algorithm applied to "iris_setosa_versicolor" using 60% training data and 40% test data.

Similarly, the contour diagram below shows the decision regions separated by a decision boundary (zero contour height). This decision boundary is indicated by the greenish-bluish line labelled with '0'.

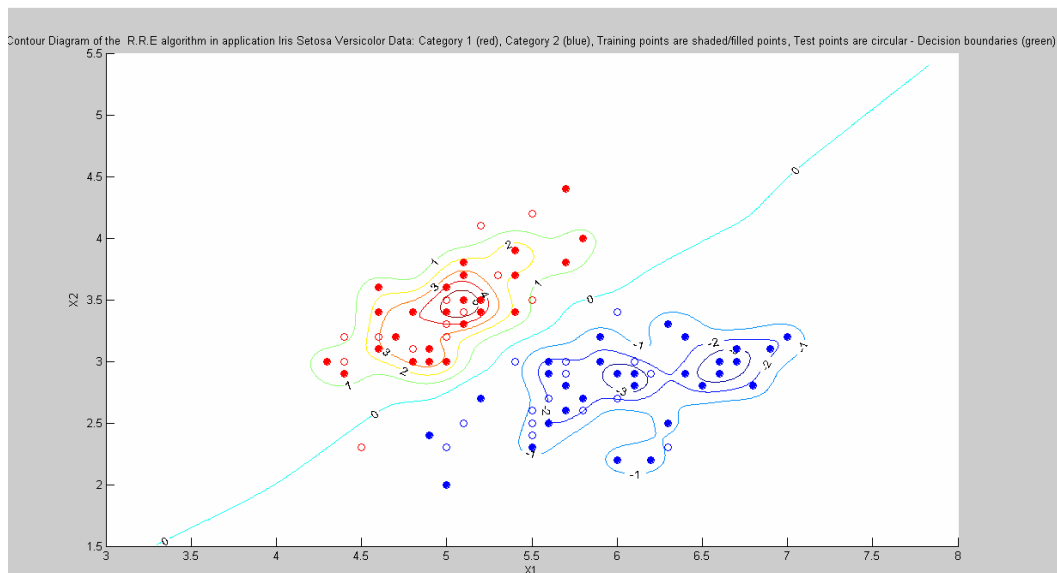

**Figure 3.2.b2:** Contour Diagram of R.R.E algorithm applied to "iris_setosa_versicolor" using 60% training data and 40% test data.
Note: decision boundary indicated by '0' (zero contour height). We have a parasitic outlier at $(4.5, 2.3)^T$ that is solely misclassified.

### 3.2(c): 90% Training data and 10% test data

The dataset of ("iris_setosa_versicolor") was split into two sets with the first 90% of the data in one set(training set) and the remaining 10%(test set) in the second set. Repeating section 3.2(b) with 90% training data and 10% test data, we obtain a classification summarized by the table below.

**TABLE 3.2.c**: Classification summary of R.R.E algorithm applied to "iris_setosa_versicolor" using 90% training data and 10% test data.

| Category | Number of training vectors Misclassified | Number of testing vectors Misclassified | Classifier Accuracy excluding Training Data | Classifier Accuracy including Training Data |
|---|---|---|---|---|
| one | 0/45 | 0/5 | 100%% | 100% |
| two | 0/45 | 0/5 | 100% | 100% |
| both | 0/90 | 0/10 | **100%** | 100% |

The decision Surface of the configuration above (using 90% training data and 10% test data) is provided below:

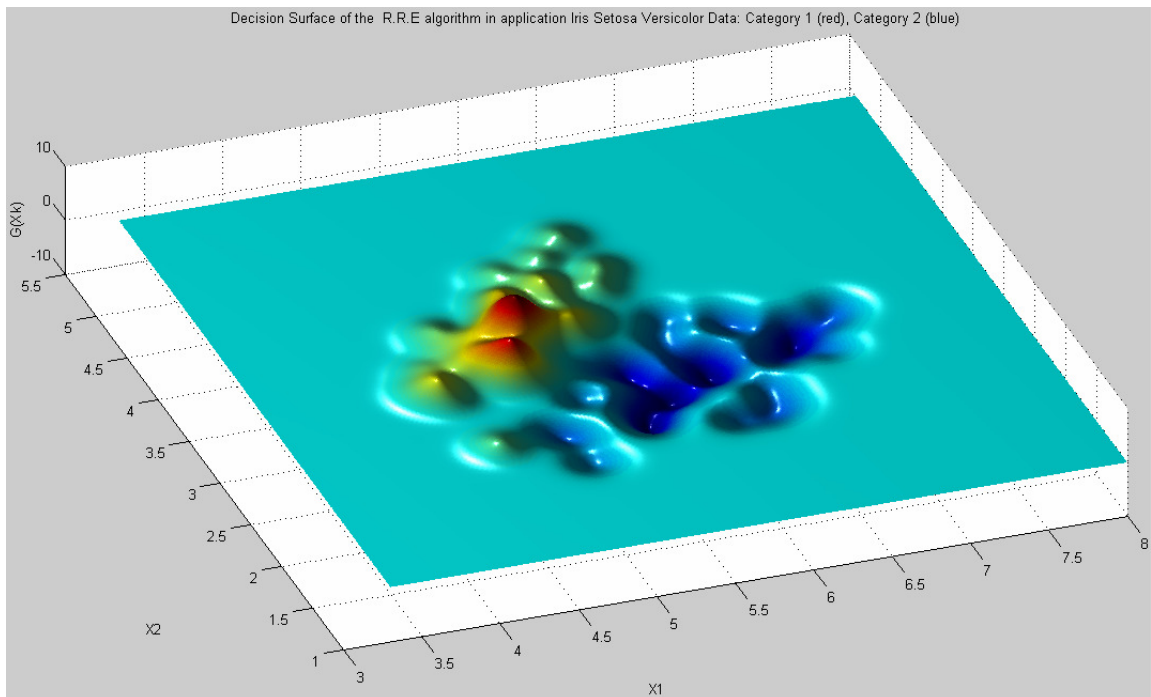

**Figure 3.2.c**: Decision surface of R.R.E algorithm applied to "iris_setosa_versicolor" using 90% training data and 10% test data.

Similarly, the contour diagram below shows the decision regions separated by a decision boundary (zero contour height). This decision boundary is indicated by the greenish-bluish line labelled with '0'.

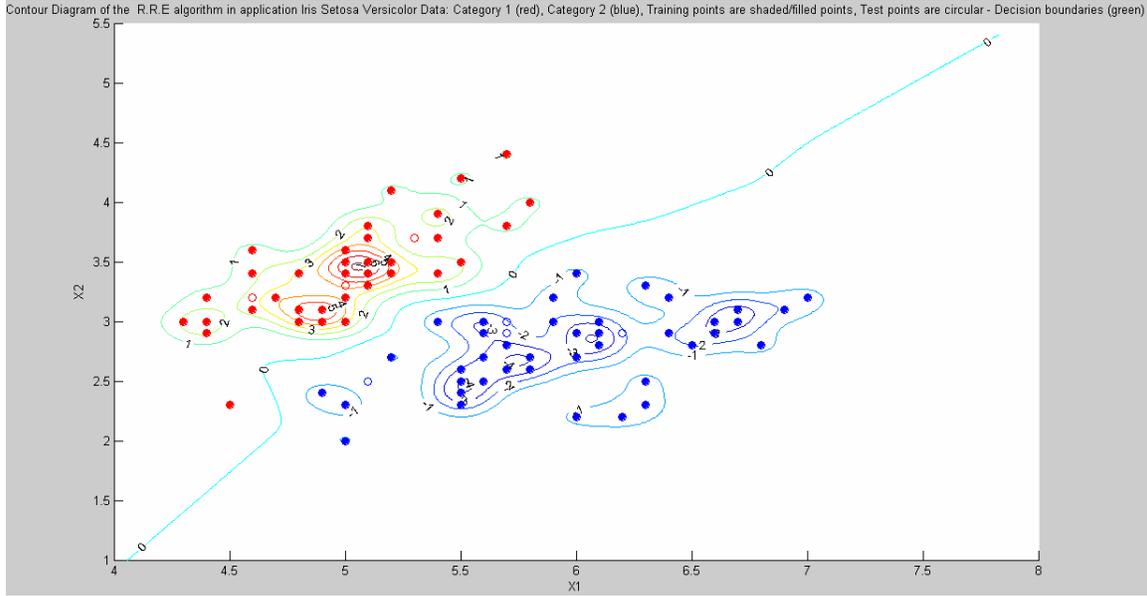

**Figure 3.2.c2**: Contour Diagram of R.R.E algorithm applied to "iris_setosa_versicolor" using 90% training data and 10% test data.
Note: decision boundary indicated by '0' (zero contour height). We have finally resolved the parasitic outlier at $(4.5,2.3)^T$ because it has now been included in training, and has influenced the decision boundary effectively.

## 3.3 Applying P.C.A to the "iris setosa versicolor" Dataset
The objective of this section is to repeat section 3.2 (3.2(a), 3.2(b), & 3.2(c)) using P.C.A.

**3.3(a)**: 40% Training data and 60% test data
The "iris_setosa_versicolor" dataset was split into two sets with 40% of the data in one set and the remaining 60% in the second set.
The first 40% of the Data set is the training data that was used to compute the weight vector $\vec{a}$. We used a learning rate, $\eta(k) = \eta(.) = 0.01$, a threshold or criterion of θ=0 and an initial weight $\vec{a}_{initial}$ = [0, 0, 1]$^T$. The number of iterations was limited to 300.
After implementing P.C.A using the abovementioned conditions, We get a final weight vector of $\vec{a}_{final} = [0.2100, -2.5860, 4.2240]^T$ by fully converging after **42** iterations. After running the test data on the final weight vector above, we get a classification summarized by the table below.

**TABLE 3.3a**: Classification summary of P.C.A applied to "iris_setosa_versicolor" using 40% training data and 60% test data.

| Category | Number of training vectors Misclassified | Number of testing vectors Misclassified | Classifier Accuracy excluding Training Data | Classifier Accuracy including Training Data |
|---|---|---|---|---|
| one | 0/20 | 2/30 | 93.33% | 96% |
| two | 0/20 | 0/30 | 100% | 100% |
| both | 0/40 | 2/60 | **96.67%** | 98% |

To capture the learning behaviour of P.C.A, the criterion function over iterations is shown below.

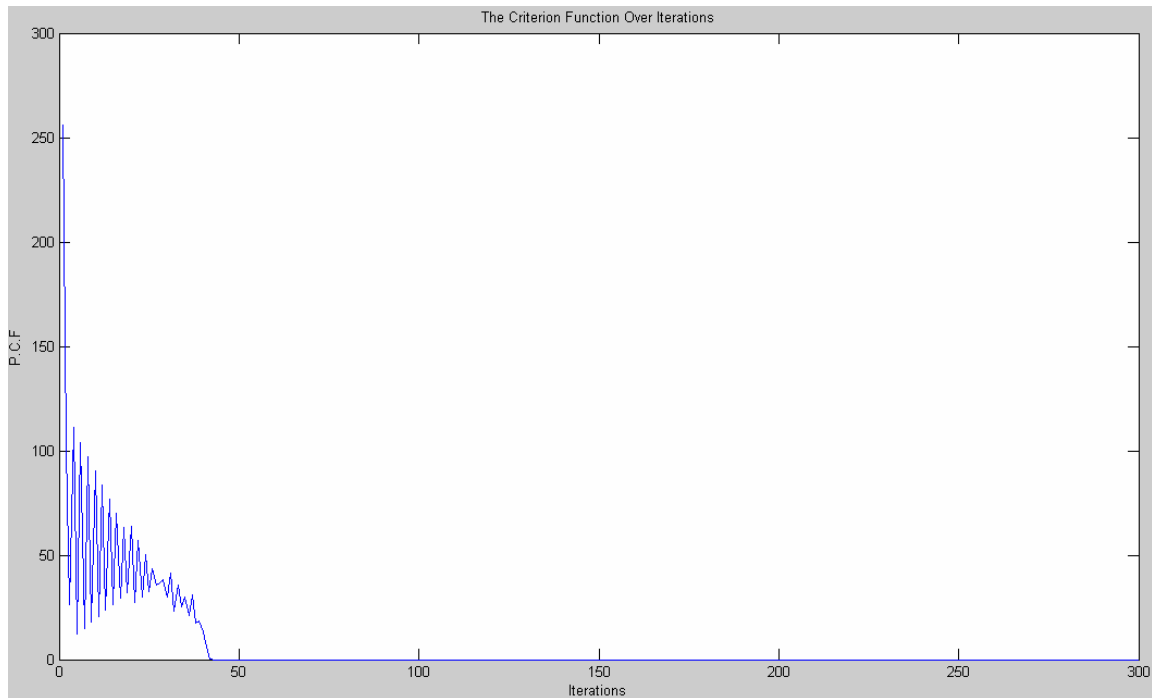

**Figure 3.3a**: Criterion function over the iterations of P.C.A applied to "iris_setosa_versicolor" using 40% training data and 60% test data.

In order to compare P.C.A to R.R.E, the Feature Space plot and the Decision Boundary is presented below.

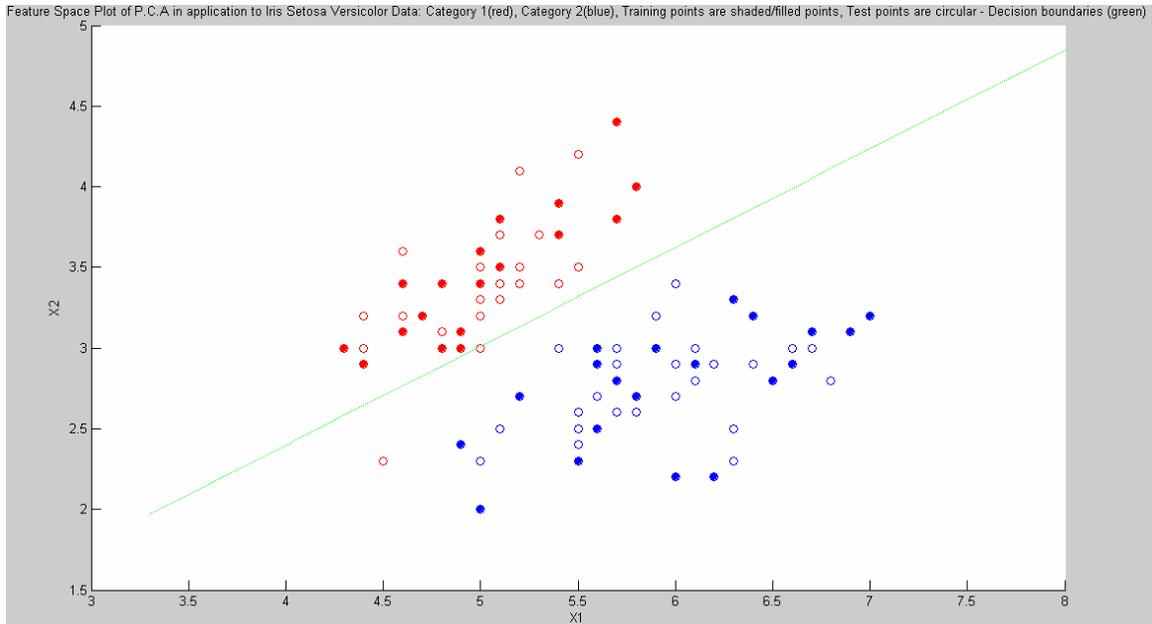

**Figure 3.3a2**: Feature Space plot and the Decision Boundary of P.C.A applied to "iris_setosa_versicolor" using 40% training data and 60% test data

### 3.3(b): 60% Training data and 40% test data

Repeating section 3.3(a) with 60% of data as training data and 40% as testing data, we get a final weight vector of $\vec{a}_{final} = [0.5100, -3.9360, 6.4120]^T$ by fully converging after **45** iterations. After running the test data on the final weight vector above, we get a classification summarized by the table below.

**TABLE 3.3.b**: Classification summary of P.C.A applied to "iris_setosa_versicolor" using 60% training data and 40% test data.

| Category | Number of training vectors Misclassified | Number of testing vectors Misclassified | Classifier Accuracy excluding Training Data | Classifier Accuracy including Training Data |
|---|---|---|---|---|
| one | 0/30 | 1/20 | 95% | 98% |
| two | 0/30 | 0/20 | 100% | 100% |
| both | 0/60 | 1/40 | **97.5%** | 99% |

Likewise, to capture the learning behaviour of P.C.A, the criterion function over iterations is shown below.

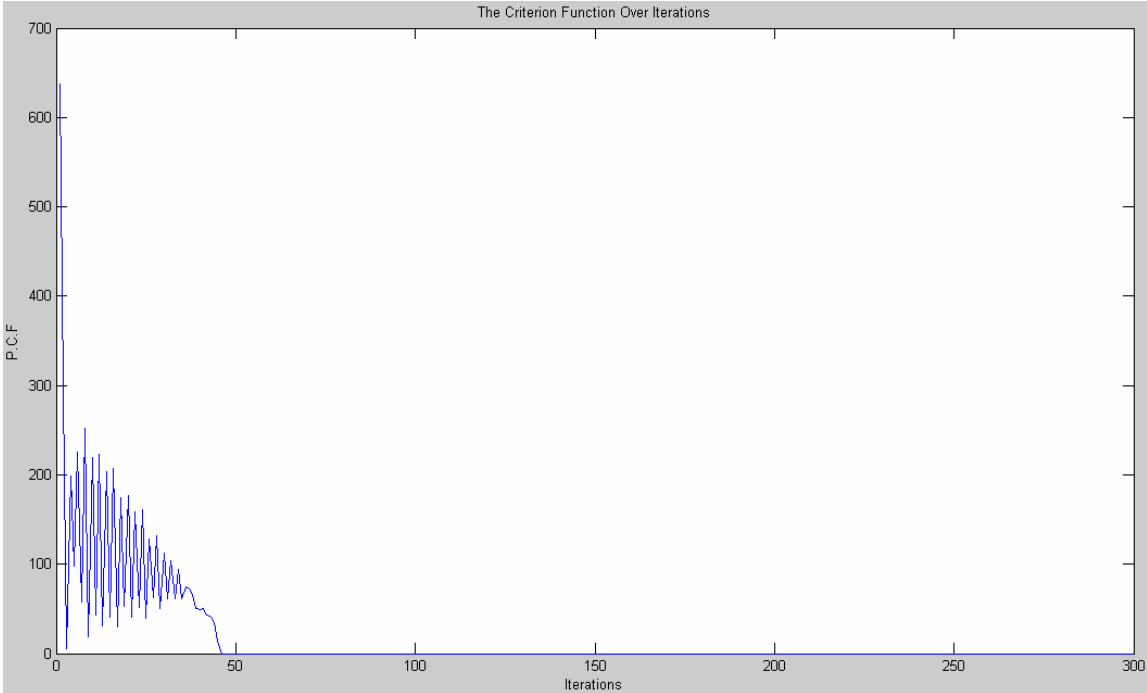

**Figure 3.3.b:** Criterion function over the iterations of P.C.A applied to "iris_setosa_versicolor" using 60% training data and 40% test data.

Again, to compare P.C.A to R.R.E, the Feature Space plot and the Decision Boundary is presented below.

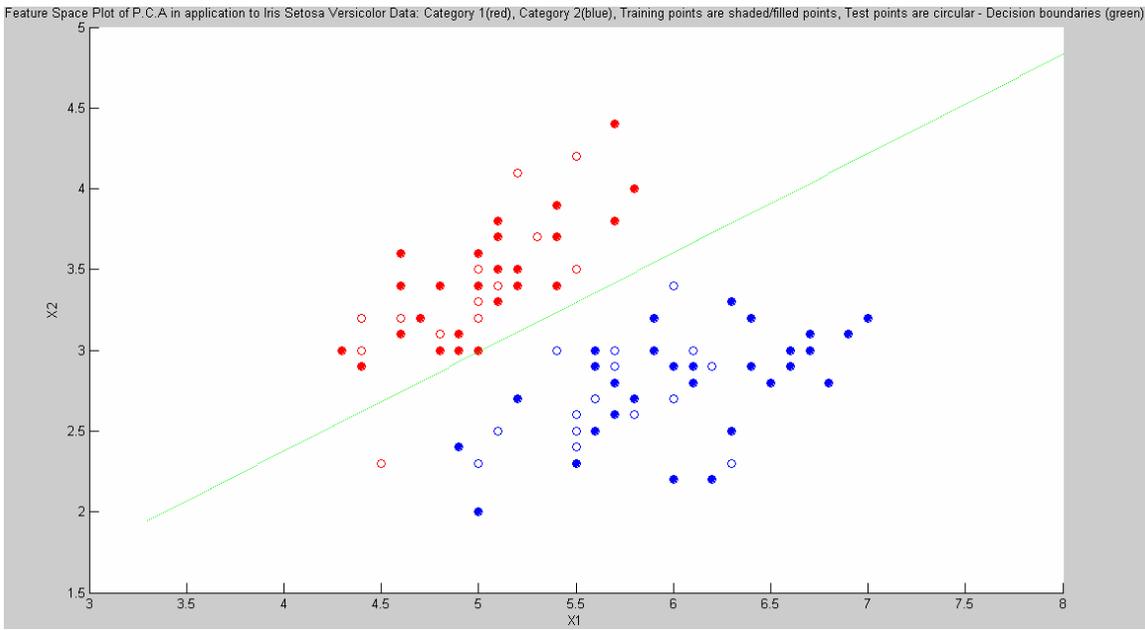

**Figure 3.3.b2:** Feature Space plot and the Decision Boundary of P.C.A applied to "iris_setosa_versicolor" using 60% training data and 40% test data

**3.3(c):** 90% Training data and 10% test data
Repeating 3.3(b) with 90% of data as training data and 10% as testing data, we get a final weight vector of
$\vec{a}_{final} = [1.4300, -5.4640, 9.2230]^T$ **without** fully converging after **300** iterations (convergence impossible due to linearly non-separable training set). After running the test data on the final weight above, we get a classification summarized by the table below.

**TABLE 3.3c:** Classification summary of P.C.A applied to "iris_setosa_versicolor" using 90% training data and 10% test data.

| Category | Number of training vectors Misclassified | Number of testing vectors Misclassified | Classifier Accuracy excluding Training Data | Classifier Accuracy including Training Data |
|---|---|---|---|---|
| one | 1/45 | 0/5 | 100% | 98% |
| two | 0/45 | 0/5 | 100% | 100% |
| both | 1/90 | 0/10 | **100%** | 99% |

Similarly, to capture the learning behaviour of P.C.A, the criterion function over iterations is shown below.

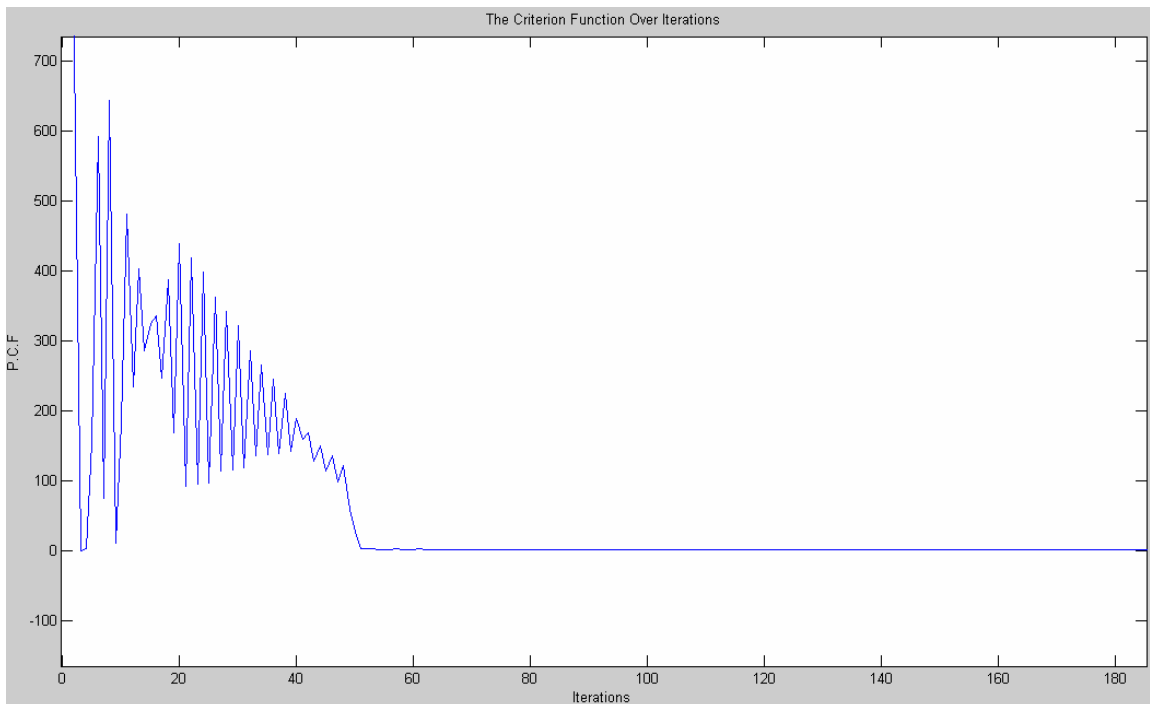

**Figure 3.3c:** Criterion function over the iterations of P.C.A applied to "iris_setosa_versicolor" using 90% training data and 10% test data.

Lastly, to compare P.C.A to R.R.E, the Feature Space plot and the Decision Boundary is presented below.

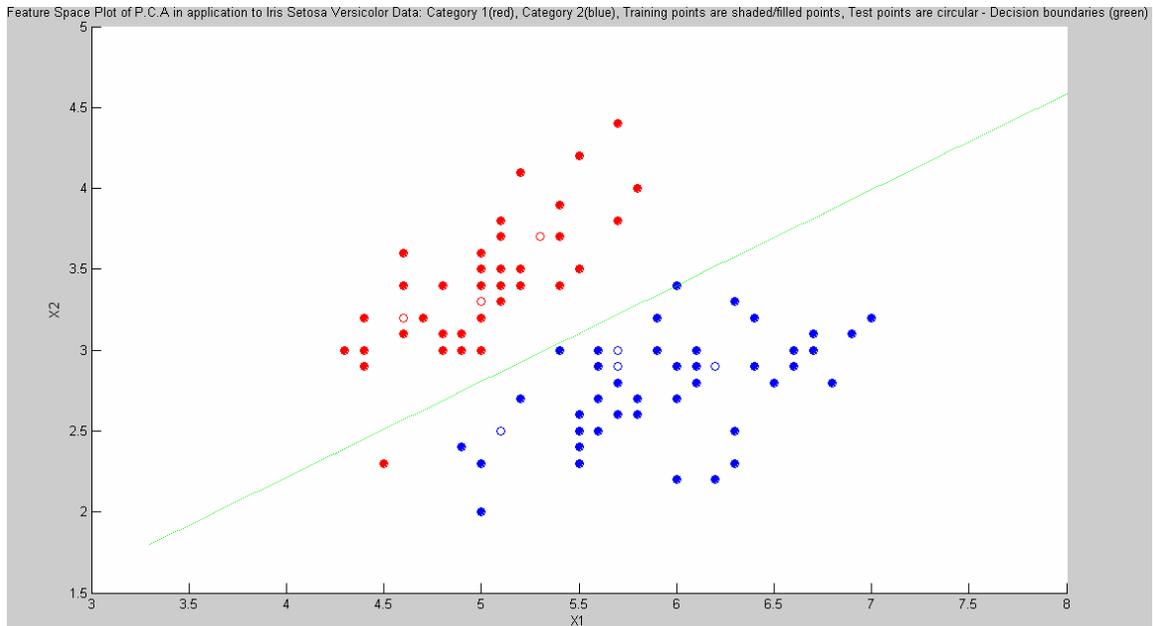

**Figure 3.3c2**: Feature Space plot and the Decision Boundary of P.C.A applied to "iris_setosa_versicolor" using 90% training data and 10% test data

## 3.4: Using "Iris Versicolor VirginicaV2" Dataset

In sections 3.1 -3.3, the R.R.E and P.C.A were implemented on a dataset that was favourable to P.C.A (because the data is somewhat linearly correlated along each category), despite this, the R.R.E achieved better overall classification results. In this section, we compare both algorithms to a dataset that is not that friendly - a dataset that is separable but not linearly separable as a convincing argument that **R.R.E can achieve 100% learning/training accuracy** while the P.C.A can not. We use the "Iris_Versicolor_VirginicaV2", a mild modification of Iris_Versicolor_Virginica" such that the 10 overlapping categorical data points have been marginally adjusted to not overlap.

The figure below visually depicts the "Iris_Versicolor_VirginicaV2" dataset. Refer to Appendix B2 to view a tabular listing of all normalized and augmented row vectors contained in the dataset. We can see the data is quite 'noisy' because data points from each category are heavily shuffled together.

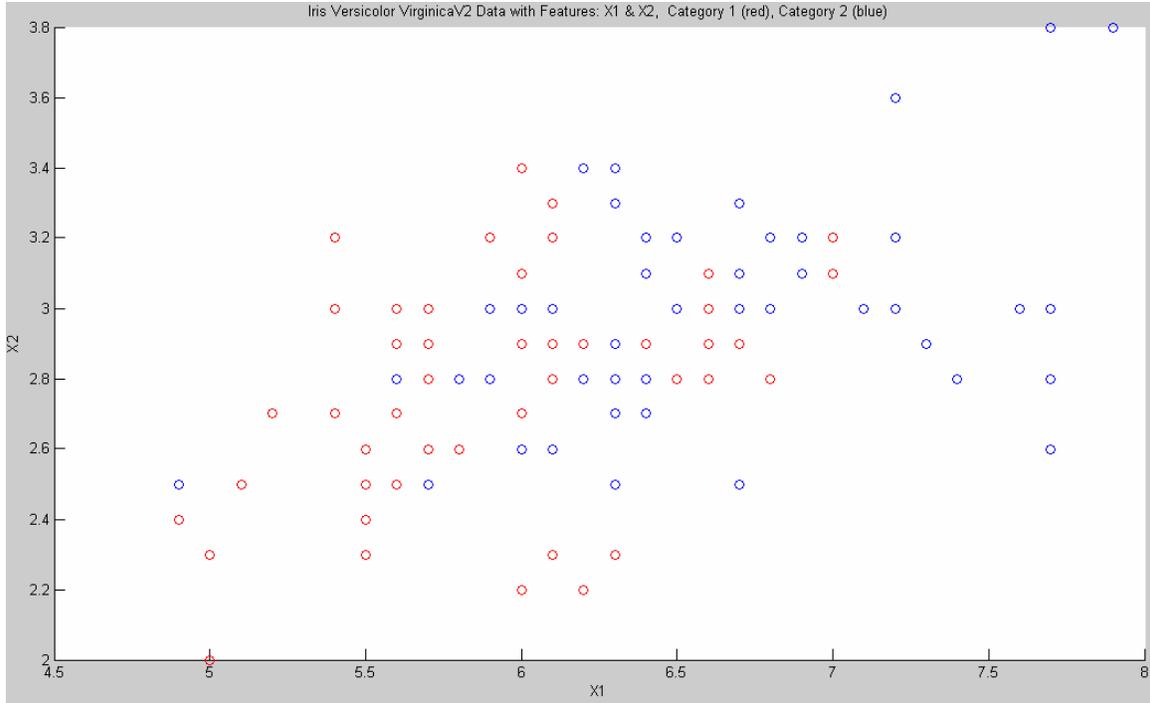
**Figure 3.4**: Visual representation of "Iris_Versicolor_VirginicaV2" dataset

## 3.5: Applying R.R.E Algorithm to the "Iris Versicolor VirginicaV2" dataset

The goal is to achieve the highest training accuracy on the entire dataset in order to assess the training ability of the algorithm.

Using our Discriminant function below,

$$G(x_k) = p_2 \sum_i e^{-\lambda f(n_i)(x_k - x_i)^T (x_k - x_i)} - p_1 \sum_j e^{-\lambda f(n_j)(x_k - x_j)^T (x_k - x_j)} \quad \forall x_i \in T_1, \forall x_j \in T_2 \text{ and } x_k \in V_k$$

Assuming equal unitary categorical cost ($p_1 = p_2$) and a variance reduction function of $f(n_i) = f(n_j) = n_i = n_j = 50$, our Discriminant function reduces to:

$$G(x_k) = \sum_i e^{-50\lambda (x_k - x_i)^T (x_k - x_i)} - \sum_j e^{-50\lambda (x_k - x_j)^T (x_k - x_j)} \quad \forall x_i \in T_1, \forall x_j \in T_2 \text{ and } x_k \in V_k$$

Empirically, an auxiliary sensitivity factor of $\lambda \geq 3.5$ allows us to achieve 100% learning/training accuracy. The decision surface of the configuration above using 100% of the dataset as training data is seen below:

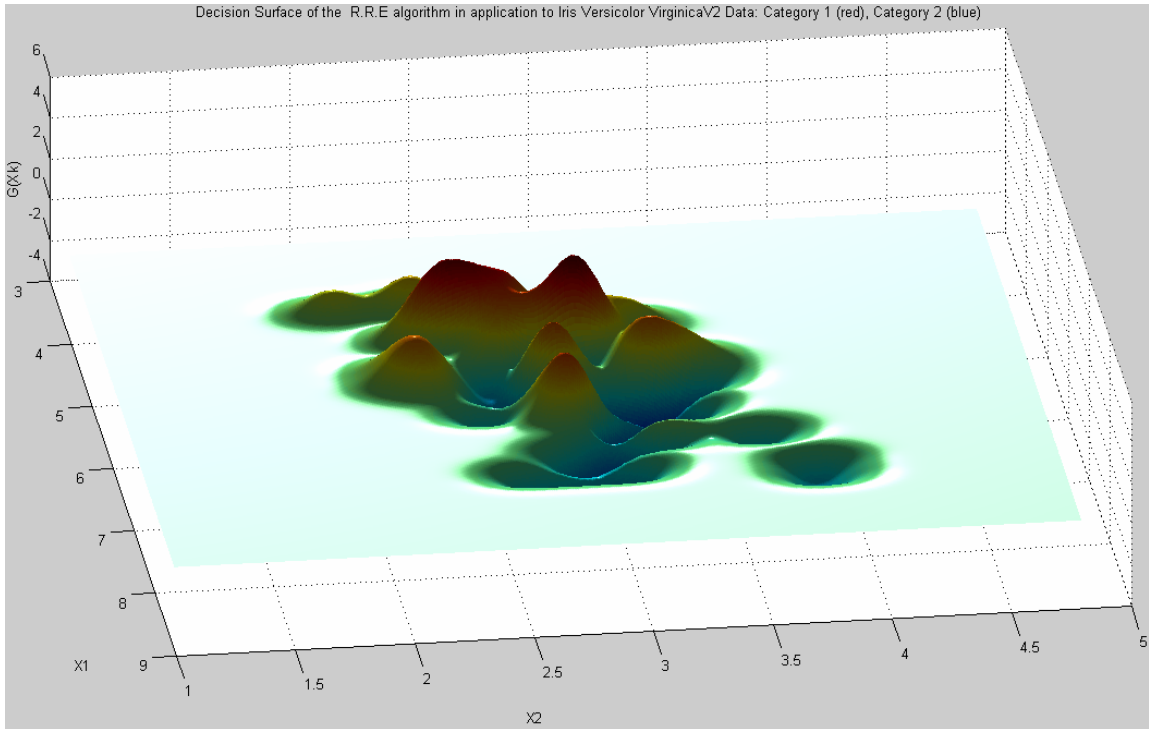

**Figure 3.5:** Decision surface of R.R.E algorithm applied to "Iris_Versicolor_VirginicaV2" using 100% training data.

The contour diagram below shows the decision regions separated by decision boundaries which are the contours of zero height (indicated by the green lines labelled '0')

Note: 100% training/learning accuracy has been achieved; it can be verified by inspecting the contours and observing that no training point is misplaced or misclassified.

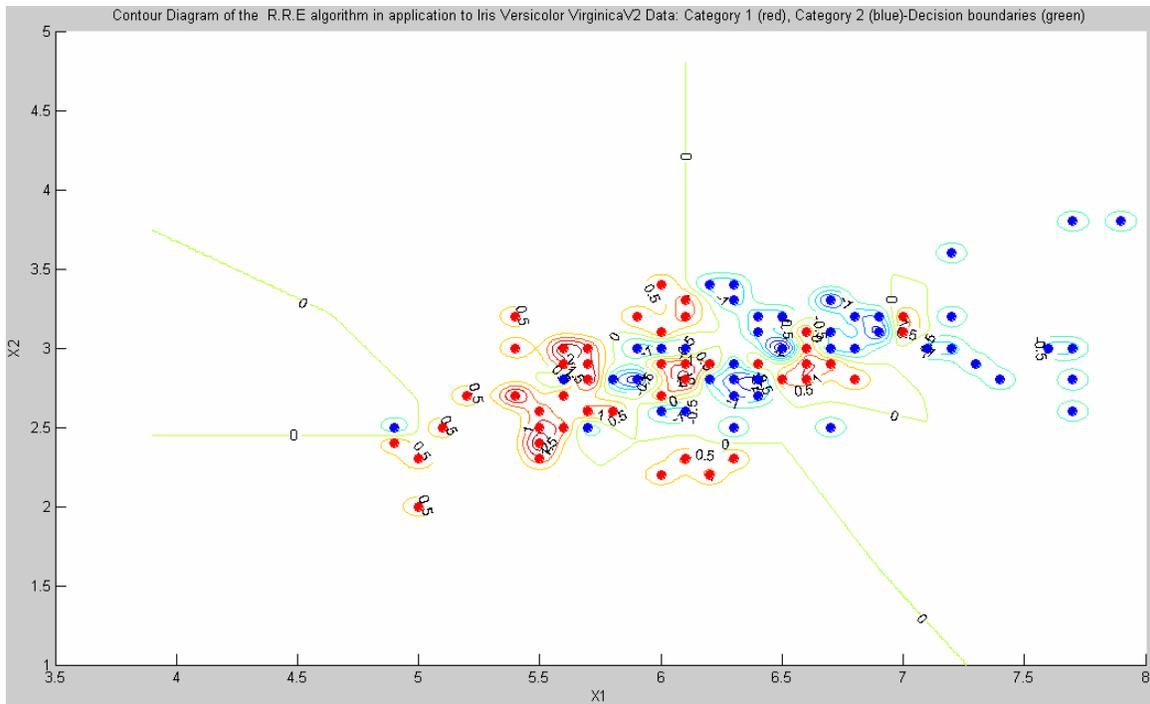

**Figure 3.5.2:** Contour Diagram of R.R.E algorithm applied to "Iris_Versicolor_VirginicaV2" using 100% training data.
Note: 100% training/learning accuracy achieved; can be verified by inspecting the contours and observing that no training point is misclassified.

## 3.6: Applying P.C.A to the "Iris VersicolorVirginicaV2" Dataset

We used 100% of the data set as our training data to compute the weight vector $\vec{a}$. We used a learning rate $\eta(k) = \eta(.) = 0.01$, a threshold or criterion of θ=0 and an initial weight $\vec{a}_{initial}$ = [0, 0, 1]$^T$.
We get a final weight vector of
$\vec{a}_{final} = [56.5, -13.228, 5.1680]^T$ without fully converging after **3000** iterations. Note that we can not converge because the data is linearly non-separable. We achieve **56% learning/training accuracy.**

To capture the learning behaviour of P.C.A under the conditions of section 3.3, the criterion function over iterations is shown below. Notice the non-convergence; the oscillatory behaviour is not because the learning rate is too high, it is due to the fact that the data is linearly non-separable.

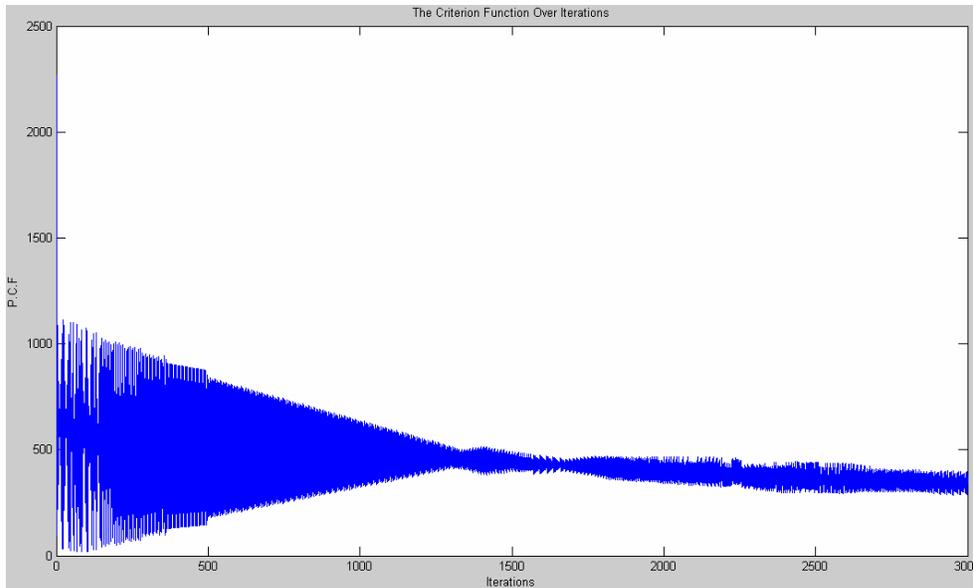
**Figure 3.6**: Criterion function over the iterations of P.C.A applied to "Iris_Versicolor_VirginicaV2 using 100% training data. No convergence due to training data being linearly non-separable

Lastly, to compare P.C.A to R.R.E, the Feature Space plot and the Decision Boundary is presented below.

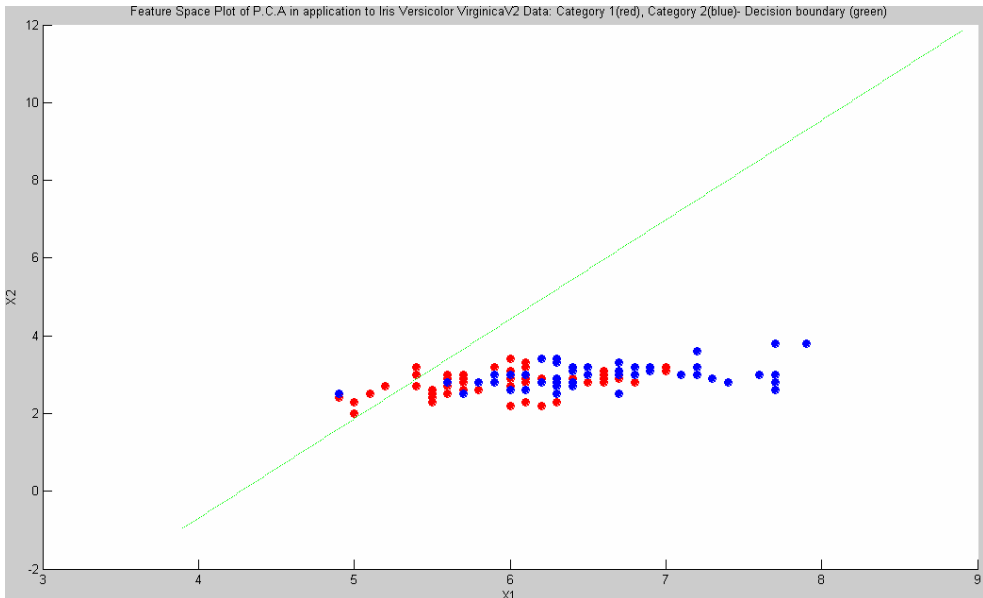
**Figure 3.6.2**: Feature Space plot and the Decision Boundary of P.C.A applied to "Iris_Versicolor_VirginicaV2" using 100% training data

### 3.7: Overall Comparison
The table below summarizes the results of sections 3.2 and 3.3 plus it includes results of extrapolated configurations (different training data to test data ratios) that were not mentioned.

**TABLE 3.7**: Classification comparison between R.R.E and P.C.A using "iris_setosa_versicolor" dataset for various training to test data ratios.

| Ratio of Training data to test data | Algorithm | Number of Training Vectors Misclassified | Number of Testing Vectors Misclassified | Classifier Accuracy Excluding Training Data | Classifier Accuracy Including Training Data |
|---|---|---|---|---|---|
| 1:9 | R.R.E | 0/10 | 5/90 | 94.44% | 95% |
|  | P.C.A | 0/10 | 1/90 | 98.89% | 99% |
| **2:8** | **R.R.E** | **0/20** | **1/80** | **98.75%** | **99%** |
|  | **P.C.A** | **0/20** | **1/80** | **98.75%** | **99%** |
| 3:7 | R.R.E | 0/30 | 1/70 | 98.57% | 99% |
|  | P.C.A | 0/30 | 1/70 | 98.57% | 99% |
| **4:6** | **R.R.E** | **0/40** | **1/60** | **98.33%** | **99%** |
|  | **P.C.A** | **0/40** | **2/60** | **96.67%** | **98%** |
| 5:5 | R.R.E | 0/50 | 1/50 | 98.00% | 99% |
|  | P.C.A | 0/50 | 2/50 | 96.00% | 98% |
| **6:4** | **R.R.E** | **0/60** | **1/40** | **97.50%** | **99%** |
|  | **P.C.A** | **0/60** | **1/40** | **97.50%** | **99%** |
| 7:3 | R.R.E | 0/70 | 1/30 | 96.67% | 99% |
|  | P.C.A | 0/70 | 1/30 | 96.67% | 99% |
| **8:2** | **R.R.E** | **0/80** | **1/20** | **95.00%** | **99%** |
|  | **P.C.A** | **0/80** | **1/20** | **95.00%** | **99%** |
| 9:1 | R.R.E | 0/90 | 0/10 | 100.0% | 100% |
|  | P.C.A | 1/90 | 0/10 | 100.0% | 99% |
| All tests above used the "iris_setosa_versicolor" dataset ||||||

The table below summarizes the results of sections 3.5 and 3.6 where we showed that when the dataset is separable, but **not** linearly separable, R.R.E can achieve 100% learning/training accuracy while the P.C.A can not.

**TABLE 3.7.2**: Training accuracy comparison between R.R.E and P.C.A using entire "Iris_Versicolor_VirginicaV2" dataset for training.

|  | Algorithm | Training Accuracy |
|---|---|---|
| **100% Training Data** | **R.R.E** | 100% |
|  | **P.C.A** | 56% |

Overall, we see that P.C.A requires a training of n iterations to adjust the augmented weight vector with a training computational complexity of $O(n)$; Since R.R.E is non-iterative, it requires minimal training in its construction of its Discriminant function with a computational complexity $O(1)$.

Once training is complete, in order to classify a single test point, P.C.A achieves classification with a computational complexity of $O(1)$- a single dot product operation, while R.R.E has a complicated Discriminant function that has an expensive computational complexity of $O(n_t)$ where $n_t$ is the number of training points.

If data is linearly non-separable, P.C.A will achieve a training accuracy of less than 100% but R.R.E seems to achieve 100% training/learning accuracy as seen in section 3.
One last merit to consider is that even if the data is linearly separable, P.C.A achieves a linear Discriminant that is not optimum with respect to the margin of separation between the support vectors of each category, on the other hand, R.R.E achieves better if not optimum margins of separation. Overall, R.R.E is indeed superior to P.C.A in terms of achieving more elegant classification results, but P.C.A is more computationally conservative upon classification.

**4 Comparing R.R.E to Linear Support Vector Machines using "Support1" & "Support2" Datasets**

**4.1: Preliminary**
The figure below visually depicts the "Support1" dataset. This dataset is a modified version of "iris_setosa_versicolor" dataset with an outlier removed such that the data is linearly separable so as to correctly implement the linear support vector machine in the next section. Refer to Appendix B3 to view a tabular listing of all normalized and augmented row vectors contained in the dataset. We can see that the dataset below is clearly linearly separable.

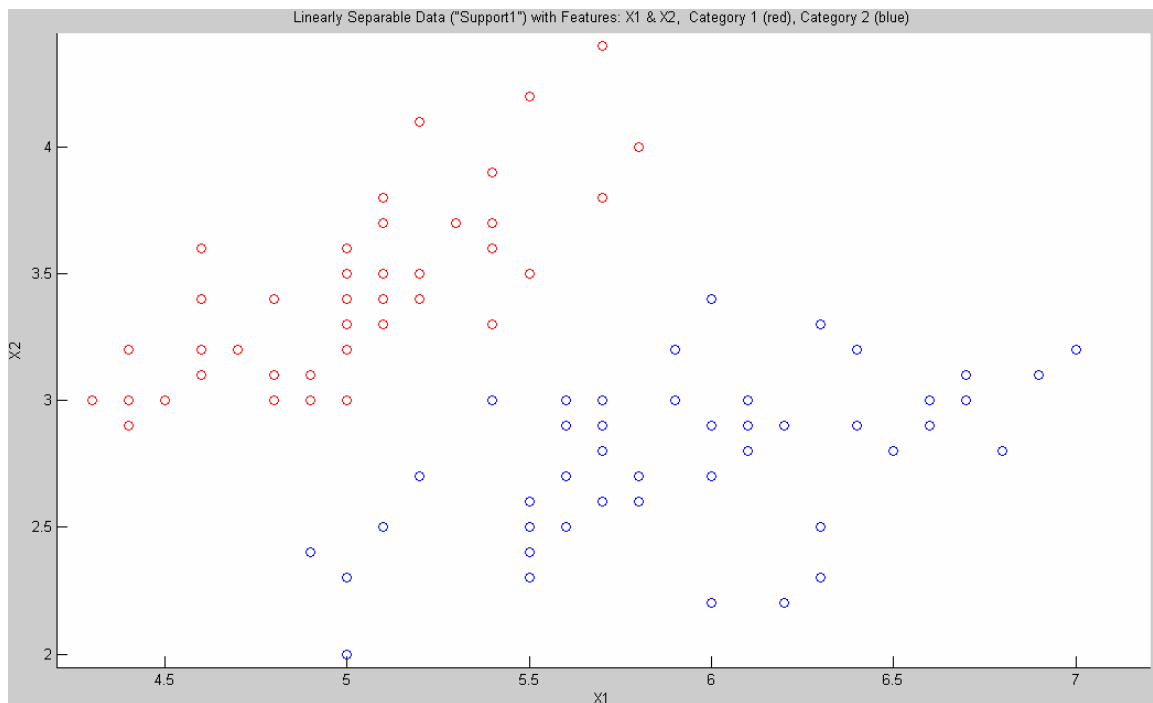

**Figure 4.1:** Visual representation of "Support1" dataset

## 4.2: Applying R.R.E Algorithm to the "Support1" Dataset

The goal of this section is to achieve optimum margin between categories by attaining a learning/training accuracy of 100%. Using our Discriminant function below,

$$G(x_k) = p_2 \sum_i e^{-\lambda f(n_i)(x_k-x_i)^T(x_k-x_i)} - p_1 \sum_j e^{-\lambda f(n_j)(x_k-x_j)^T(x_k-x_j)} \quad \forall x_i \in T_1, \forall x_j \in T_2 \text{ and } x_k \in V_k$$

Assuming equal unitary categorical cost ($p_1 = p_2$) and a variance reduction function of $\boxed{f(n_i) = f(n_j) = n_i = n_j = 50}$, our Discriminant function reduces to:

$$G(x_k) = \sum_i e^{-50\lambda (x_k-x_i)^T(x_k-x_i)} - \sum_j e^{-50\lambda (x_k-x_j)^T(x_k-x_j)} \quad \forall x_i \in T_1, \forall x_j \in T_2 \text{ and } x_k \in V_k$$

Empirically, an auxiliary sensitivity factor of $\lambda \geq 1$ allows us to achieve **100% learning/training accuracy** with an optimum non-linear margin.

The contour diagram below shows the decision regions separated by a decision boundary which is the contour height of zero (indicated by the bluish-green line labelled '0'). This decision boundary separates both categories by not only considering the linear support vectors, but rather all data. We can see an optimum margin emerging that is justifiably non-linear.

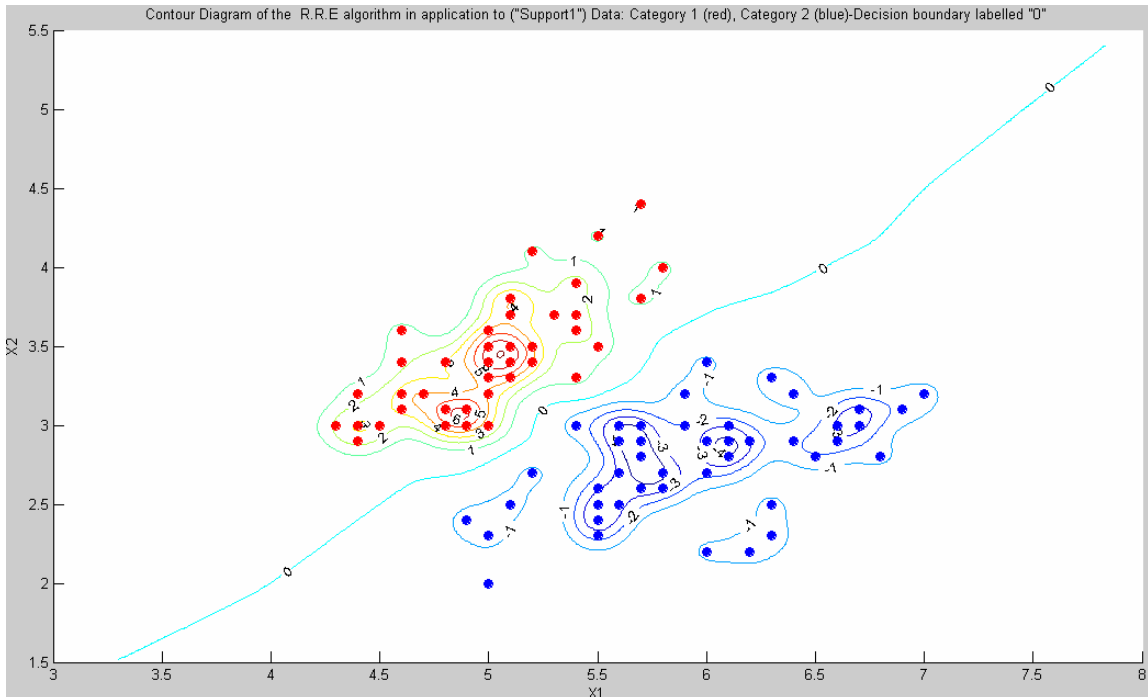

**Figure 4.2:** Contour Diagram of R.R.E algorithm applied to "Support1" dataset using 100% training data. Notice the optimum non-linear margin between support vectors.

## 4.3: Applying Linear S.V.M to the "Support1" Dataset

**Step 1:** Identifying the support vectors
In order to use the Linear S.V.M., we would first need to identify the support vectors. Assuming P.C.A was pre-implemented and the closest data points to the linear Discriminant line were obtained, we find the support vectors shown in the diagram below:

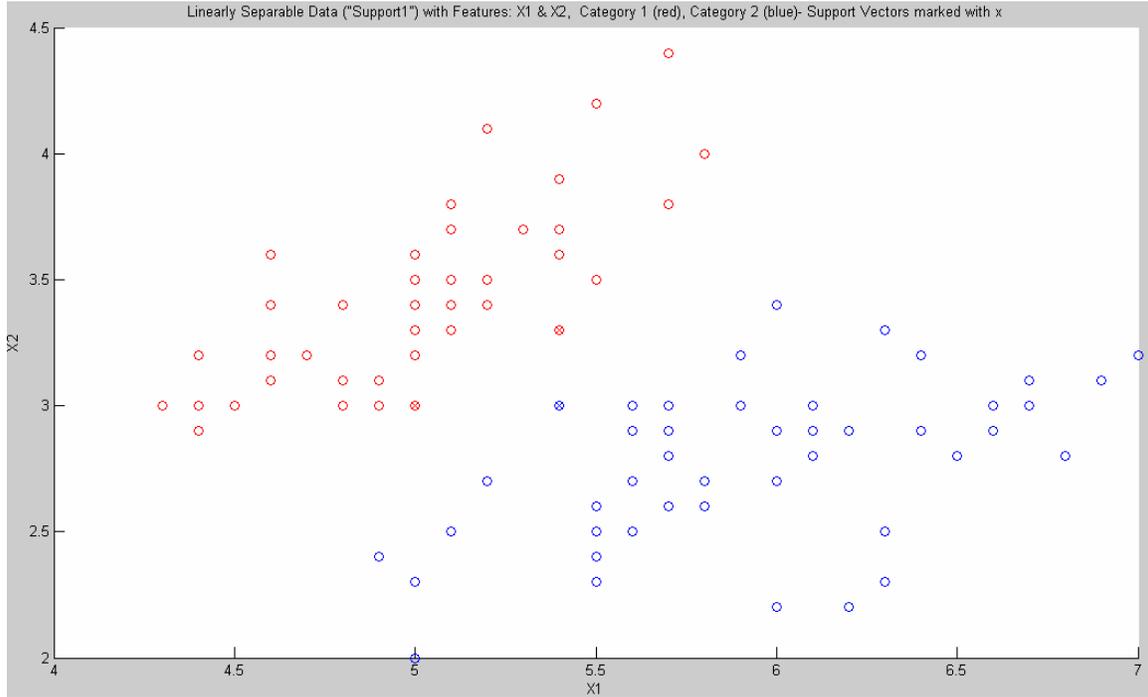

**Figure 4.3**: "Support1" dataset with indicated support vectors

**Step 2:** Augmenting and normalizing support vectors
Our support vectors from category one are: $V_{1,1}=[5,3]^T$ and $V_{1,2}=[5.4,3.3]^T$
Our support vector from category two is: $V_{2,1}=[5.4,3]^T$
After augmenting and normalizing our support vectors, we get:
$\phi_1=[1.0, 5.0, 3.0]^T$, $\phi_2=[1.0, 5.4, 3.3]^T$ and $\phi_3=[1.0, -5.4, -3.0]^T$

**Step 3:** Constructing the normalized kernel matrix
Our normalized kernel matrix is $K_{i,j}=\phi_i.\phi_j^T \quad i,j \in \{1,2,3\}$
Using MATLAB instructions:
```
q =[1,5,3,;1,5.4,3.3;-1,-5.4,-3]
k=q*transpose(q)
```
we get:
$$K_{i,j}=\begin{bmatrix} 35.0 & 37.9 & -37.0 \\ 37.9 & 41.05 & -40.06 \\ -37.0 & -40.06 & 39.16 \end{bmatrix}$$

**Step 4**: Finding Lagrangian coefficients

We need to solve for the Lagrangian coefficients $\alpha_j$ in

$$K_{i,j}\alpha_j = 1_i$$

$$\Rightarrow \begin{bmatrix} 35.0 & 37.9 & -37.0 \\ 37.9 & 41.05 & -40.06 \\ -37.0 & -40.06 & 39.16 \end{bmatrix} * \begin{bmatrix} \alpha_1 \\ \alpha_2 \\ \alpha_3 \end{bmatrix} = \begin{bmatrix} 1 \\ 1 \\ 1 \end{bmatrix}$$

$$\Rightarrow \begin{bmatrix} \alpha_1 \\ \alpha_2 \\ \alpha_3 \end{bmatrix} = \begin{bmatrix} 35.0 & 37.9 & -37.0 \\ 37.9 & 41.05 & -40.06 \\ -37.0 & -40.06 & 39.16 \end{bmatrix}^{-1} * \begin{bmatrix} 1 \\ 1 \\ 1 \end{bmatrix}$$

We find a solution to our Lagrange coefficients: $\alpha_1$, $\alpha_2$ and $\alpha_3$

$$\begin{bmatrix} \alpha_1 \\ \alpha_2 \\ \alpha_3 \end{bmatrix} = \begin{bmatrix} 93.5 \\ -37.78 \\ 49.72 \end{bmatrix}$$

**Step 5**: computing the optimum augmented weight vector

Our final augmented weight vector is $\hat{a}$

$$\hat{a} = \sum_{i=1}^{3}\alpha_i\phi_i = [6.0, -5, 20/3]^{\mathrm{T}}$$

**Step 6**: Constructing the optimum linear Discriminant function
Our Discriminant function becomes:
G(x) = $\hat{a}$.[1,x$_1$,x$_2$]$^{\mathrm{T}}$= 6 -5x$_1$ + (20/3)x$_2$ =0
Which is equivalent to:  **x$_2$ = (3/4)x$_1$ − 9/10**

**Step 7**: Confirming the solution
In order to confirm the solution of our optimum linear
Discriminant function, we find the distance or margin of all
support vectors to the optimum Discriminant line is:

$$\boxed{\rho = (\hat{a}*\phi_i^T)/(\sqrt{\hat{a}_2^{\,2} + \hat{a}_3^{\,2}}) = \mathbf{0.12}, \text{ for } i = 1,2,3}$$

Since the margins of each support vectors to the optimum
Discriminant line are equal, we have confirmed the correct
solution.

The figure below summarizes the results of steps 1 through 7 by
depicting the Feature Space plot and the optimum linear decision
Boundary.

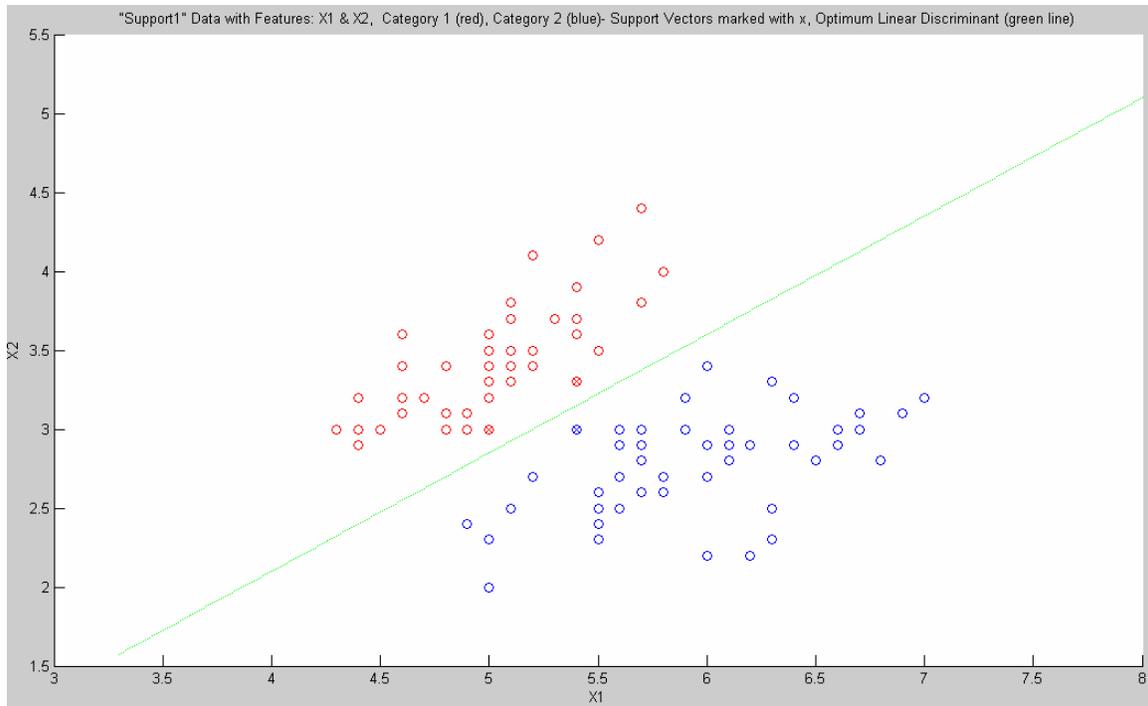

**Figure 4.3**: Feature Space plot & the optimum **linear** decision Boundary of linear S.V.M applied to "Suppot1" dataset using the indicated support vectors.

### 4.4: Applying R.R.E Algorithm to the "Support2" Dataset Versus L.S.V.M & Fisher Linear Discriminant

In section 4.3, we used a dataset ("Support1") that conveniently allowed us to apply linear S.V.M, now we draw our attention to a data scenario that compromises the functionality of both the Fisher Linear Discriminant and linear S.V.M.
We apply R.R.E under the conditions presented in 4.2.
The "support2'data was constructed from scratch: A visual representation of this data is seen below:

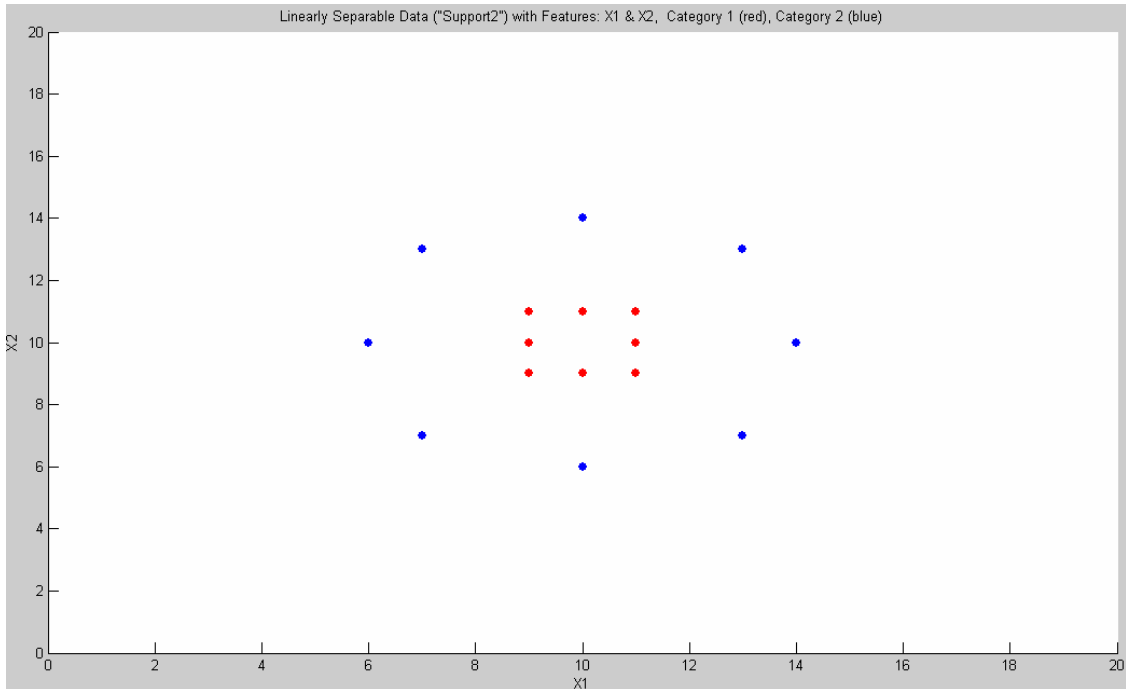
**Figure 4.4**: Visual representation of "Support2" dataset

R.R.E generates a symmetric decision surface below:

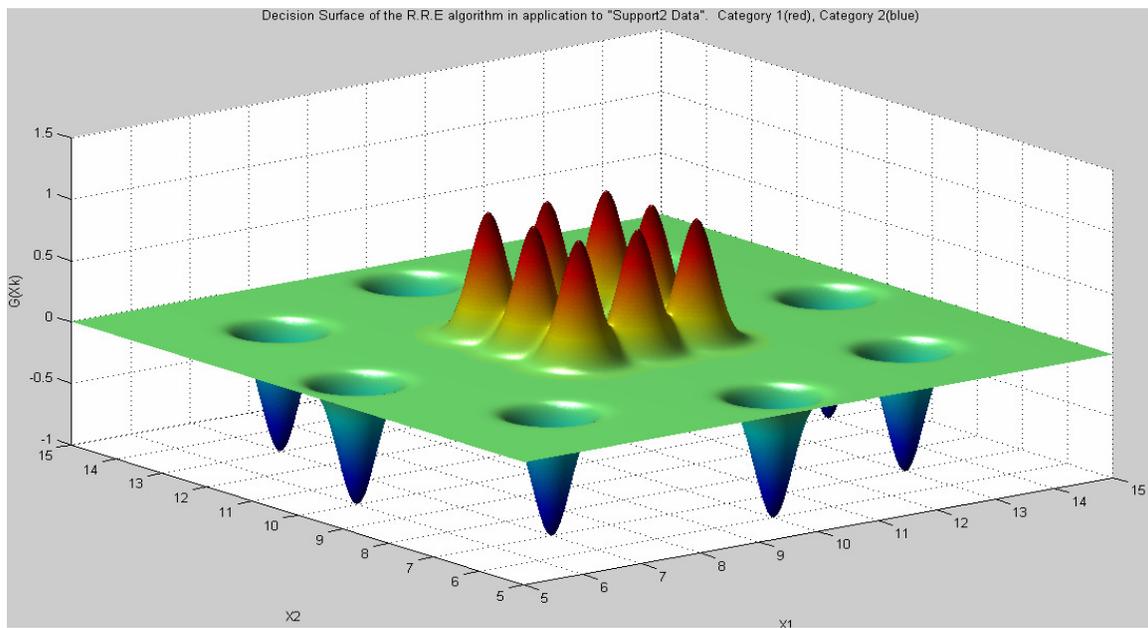
**Figure 4.4.2**: Decision surface of R.R.E algorithm applied to "Support2" dataset using 100% training data.

After implementing R.R.E, we obtain the contour diagrams of the feature space presented below:

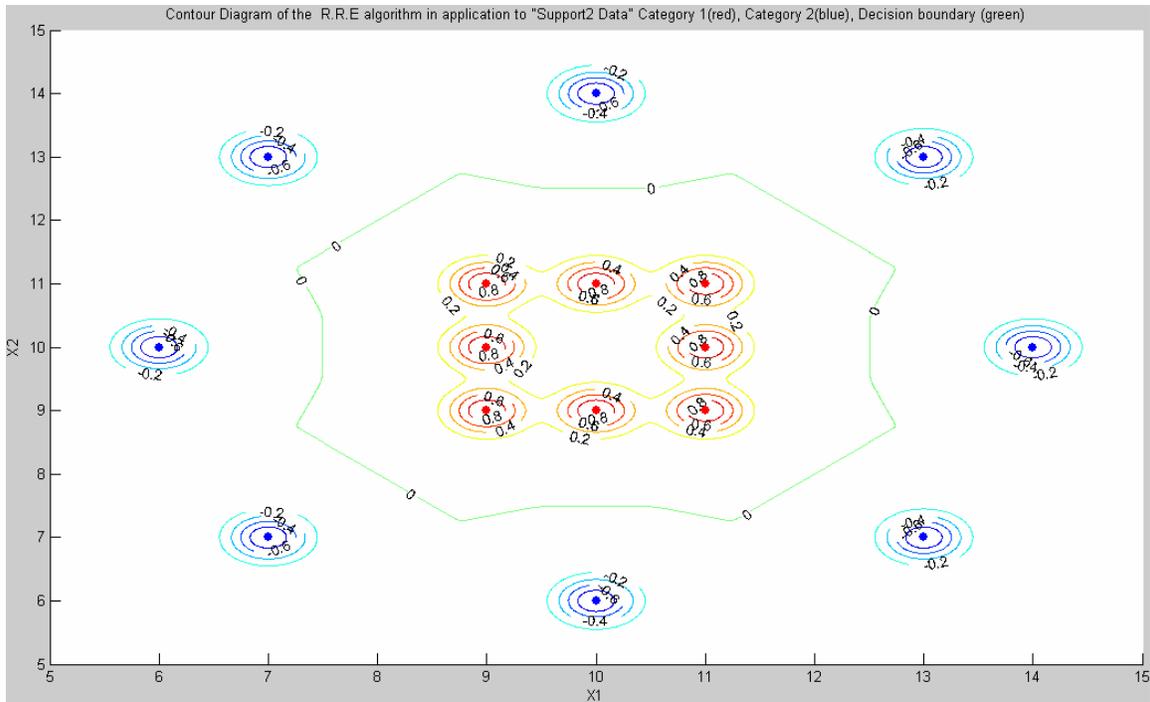

**Figure 4.4.3**: Decision surface of R.R.E algorithm applied to entire "Support2" dataset. Decision boundary marked with a contour value of '0'

The result above demonstrates how powerful R.R.E is because it is able to produce both linear and non-linear optimum margins. If a Linear S.V.M is applied to "suppot2" data, it would produce inadequate results because the data is not linearly separable. The Fisher Linear Discriminant would fail as well because the means of each category are equivalent ($\mu_1 = \mu_2 = [10,10]^T$).

Note: Non-linear S.V.M could potentially produce decent results as long us we initially project our data to some higher dimension as a preliminary step to separate data.

## 5 Comparing R.R.E to a 2-2-1 Neural Network in application to the popular XOR Problem

The popular XOR problem is a classic problem that can not be solved by linear classifiers/Perceptrons. We attempt to solve the XOR problem by employing the R.R.E algorithm and then to compare results to a solution of a Neural Network of modest complexity. We are using a Neural Network as a 'benchmark' because it is conventionally used in solving linearly non-separable classification problems. We use a 2-2-1 Neural Network because it is simple, yet computationally sufficient with respect to solving the XOR problem.

## 5.1: Applying R.R.E Algorithm to the XOR problem

Our initial Discriminant Function is:

$$G(x_k) = p_2 \sum_i e^{-\lambda f(n_i)(x_k-x_i)^T(x_k-x_i)} - p_1 \sum_j e^{-\lambda f(n_j)(x_k-x_j)^T(x_k-x_j)} \quad \forall x_i \in T_1, \forall x_j \in T_2 \text{ and } x_k \in V_k$$

Assuming equal unitary categorical cost and a variance reduction function of $f(n_i) = f(n_j) = n_i = n_j$, our Discriminant function reduces to:

$$G(x_k) = \sum_i e^{-n_i(x_k-x_i)^T(x_k-x_i)} - \sum_j e^{-n_j(x_k-x_j)^T(x_k-x_j)} \quad \forall x_i \in T_1, \forall x_j \in T_2$$

where $T_1 = \left\{ \begin{pmatrix} -1 \\ 1 \end{pmatrix}, \begin{pmatrix} 1 \\ -1 \end{pmatrix} \right\}$, $T_2 = \left\{ \begin{pmatrix} -1 \\ -1 \end{pmatrix}, \begin{pmatrix} 1 \\ 1 \end{pmatrix} \right\}$ $n_i = n_j = 2$ and $\lambda = 1$

Using our training data as our test data, the table below summarizes the performance of the R.R.E algorithm in application to the XOR problem.

**TABLE 5.1:** Classification summary of R.R.E applied to the XOR problem

| Training Points: (X1,X2) | target | $G(x_k) = $ z(out) |
|---|---|---|
| $(-1,-1)^t$ | -1 | -0.9993 |
| $(-1,1)^t$ | 1 | 0.9993 |
| $(1,-1)^t$ | 1 | 0.9993 |
| $(1,1)^t$ | -1 | -0.9993 |
| Using redundant training (duplication) or using a higher order variance reduction function(f(n)), or for $\lambda > 1$, we can arbitrarily shift 'z(out)' to 'target', but they are not required because the current results are good enough. | | |

The decision surface of the R.R.E Discriminant function is shown below.

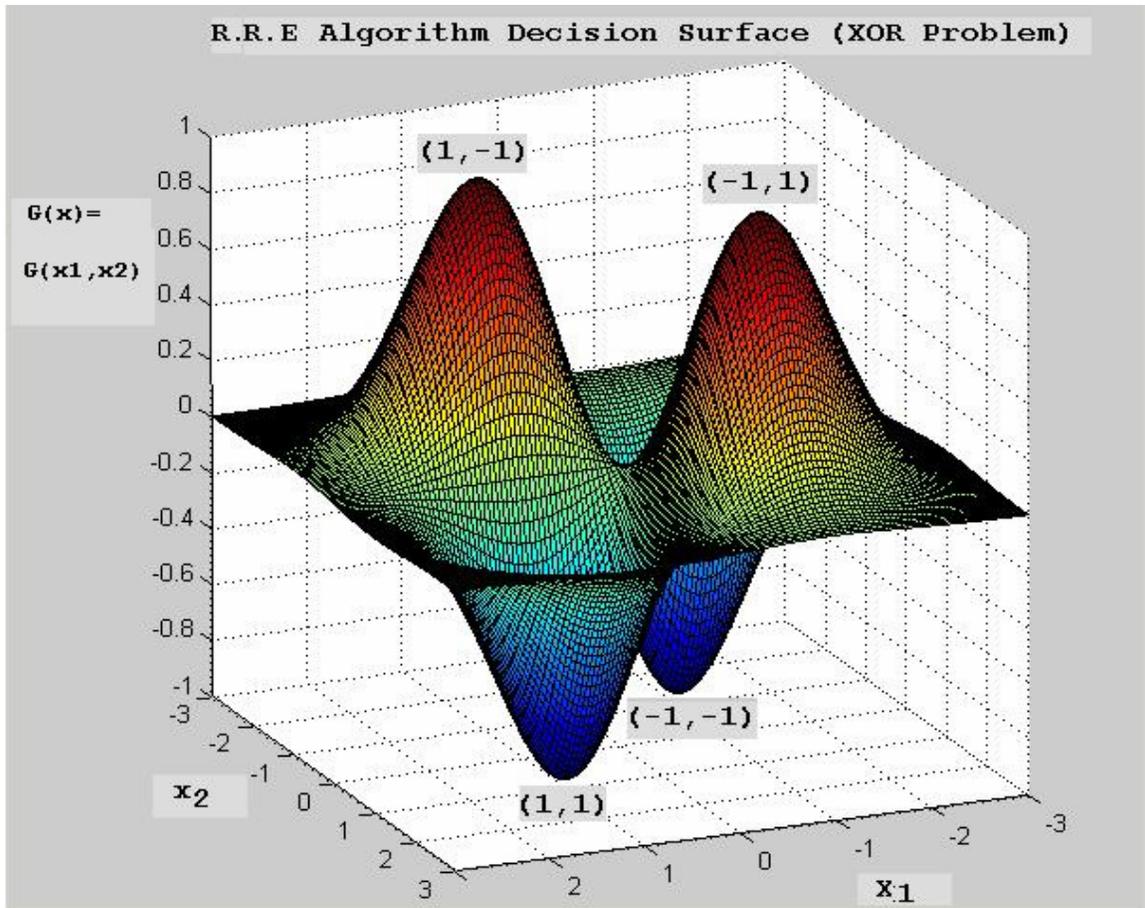
**Figure 5.1:** Mesh decision surface of R.R.E in application to the XOR problem

Projecting the surface above onto the $x_1$-$x_2$ plane, we get the following.

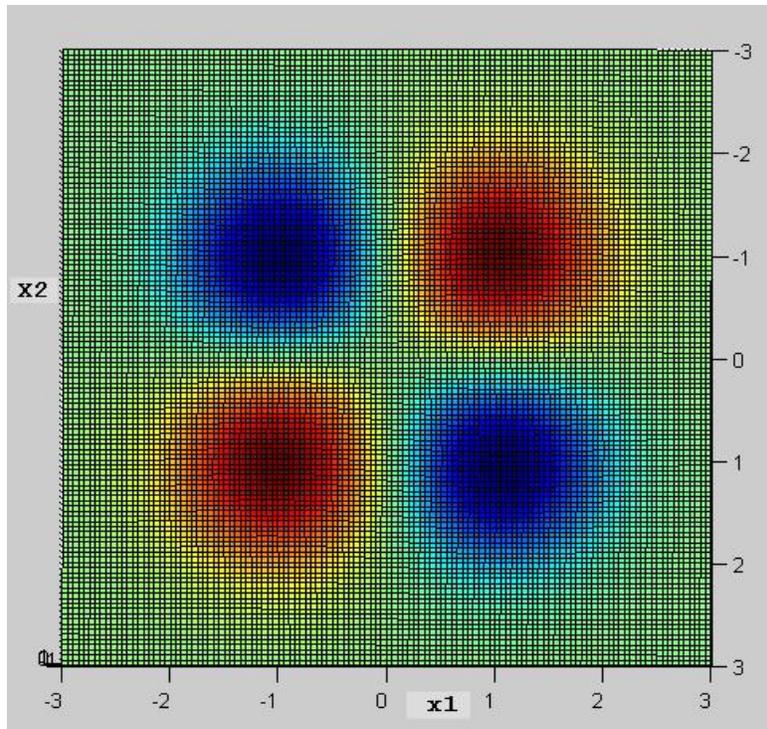

**Figure 5.1b:** projection of R.R.E decision surface onto the feature space (in application to the XOR problem)

An aesthetically alternative representation of the surface above is shown below:

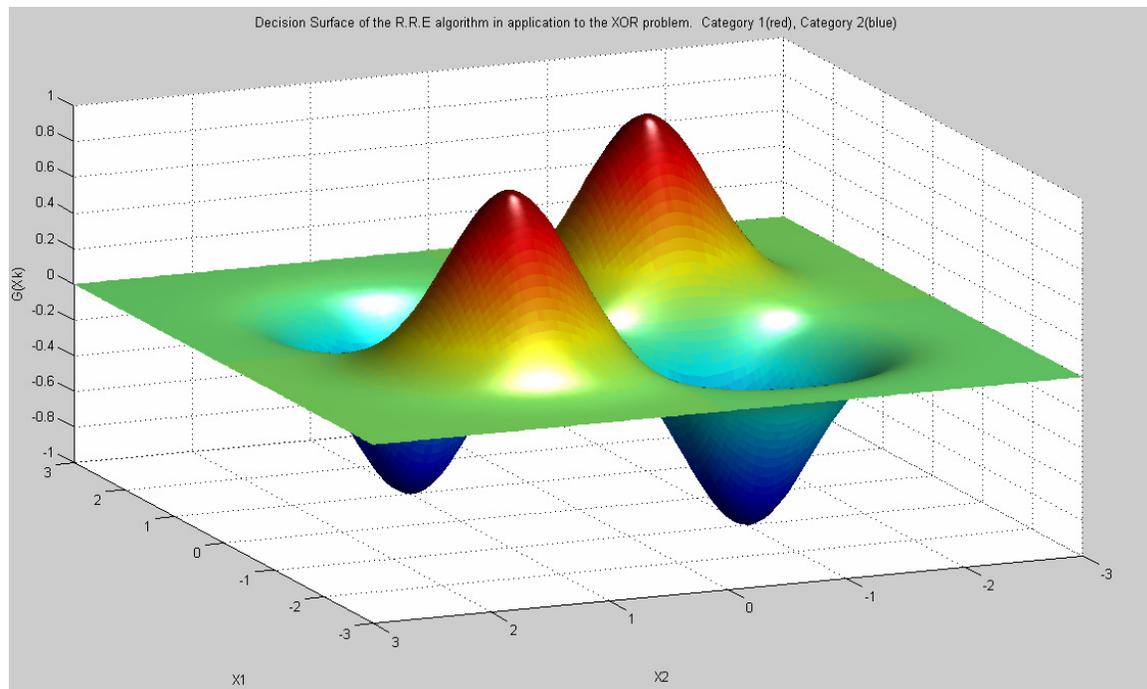

**Figure 5.1c:** Decision surface of R.R.E in application to the XOR problem

The decision surfaces shown above indeed show the behaviour of the algorithm's Discriminant function, but what is of more interest is the decision boundaries because they
visually depict the separation of decision regions. In the
diagram below, the decision boundary is seen when the
contour height is zero (straight green lines labelled '0')
which is equivalent to the Discriminant function being
zero.

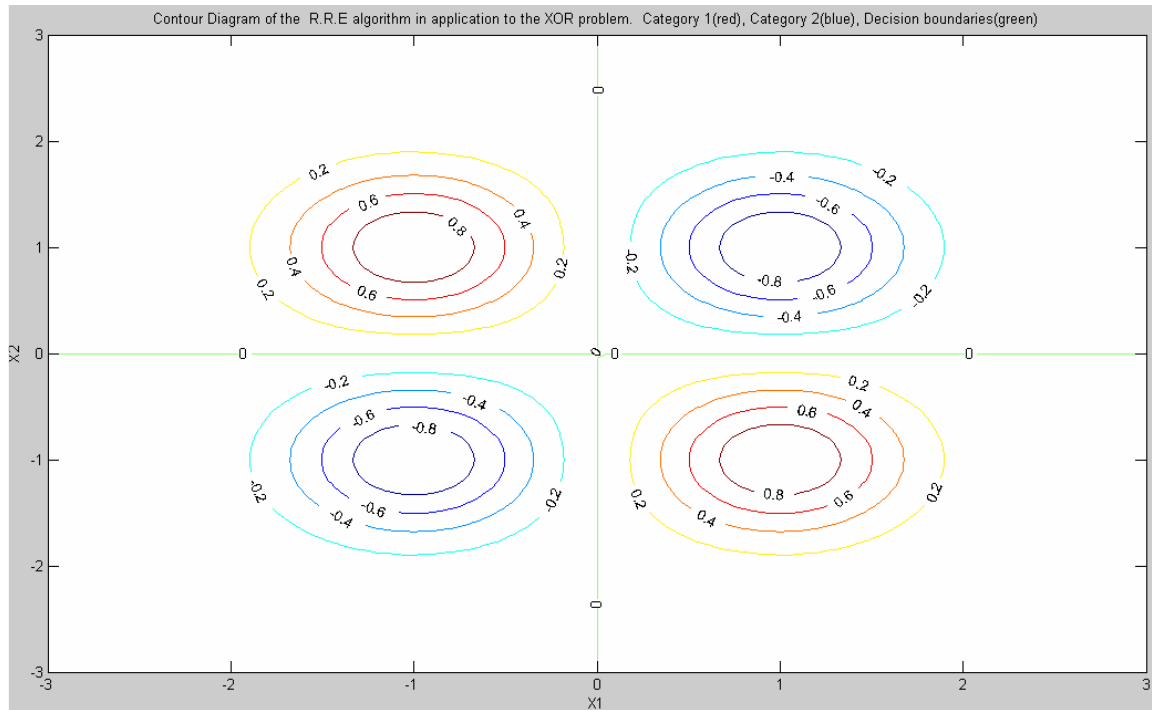
**Figure 5.1d**: Contour diagram of R.R.E in application to the XOR problem

## 5.2: Solving the XOR problem using a 2-2-1 Neural Network
A 2-2-1 Neural Network was constructed from first principles using the
batch back propagation algorithm for solving the XOR problem.

Given the following conditions:

-The two given batch inputs are:
                $x_1 = [-1,-1,1,1]$  $x_2=[-1,1,-1,1]$

- The desired (batch) target is $[-1,1,1,-1]$.
- We used a learning rate η = 0.1, a threshold (θ) of 0.001 and a
hyperbolic-tan sigmoidal activation function.

- The arbitrary chosen **initial** hidden weights are:
$W_{ji} = \begin{bmatrix} 0.1 & 0.2 \\ 0.3 & 0.4 \\ 0.\hat{5} & 0.\hat{3} \end{bmatrix}$ (Input to hidden layer weights, hats indicate bias weights)

$$W_{kj} = \begin{bmatrix} 0.27 \\ 0.31 \\ 0.\hat{2}\hat{9} \end{bmatrix}$$ (Hidden to output layer weights, hats indicate bias weights)

After running the script for the Neural Network, we verify that network converges after **32 Epochs** where the final weight values indeed satisfy the XOR operation as (**target ≈ Z(out)**).

**Table 5.2:** Classification summary of a 2-2-1 Neural Network in application to the XOR problem

| Training Points: (X1,X2) | target | Z(out) |
|---|---|---|
| $(-1,-1)^t$ | -1 | -0.9887 |
| $(-1,1)^t$ | 1 | 1.0 |
| $(1,-1)^t$ | 1 | 1.0 |
| $(1,1)^t$ | -1 | -0.9937 |

The stricter or smaller our criterion/threshold, the closer 'z(out)' comes to 'target'. **If there exists convergence & if** $\lim_{\theta \to 0}$, **then** $\lim_{Z(out) \to t \arg et}$

- Note we get a **Final Training Error** of **0.000432** < θ = 0.001
- The **Final hidden weights** after training are:

$$W_{ji} = \begin{bmatrix} 1.2031 & 3.9577 \\ 2.6005 & 4.5555 \\ -2.5788 & 5.2168 \end{bmatrix}$$ (Input to hidden layer weights)

$$W_{kj} = \begin{bmatrix} -6.4841 \\ 5.8289 \\ -3.2555 \end{bmatrix}$$ (Hidden to output layer weights)

The learning curve of the Neural Network is presented below:

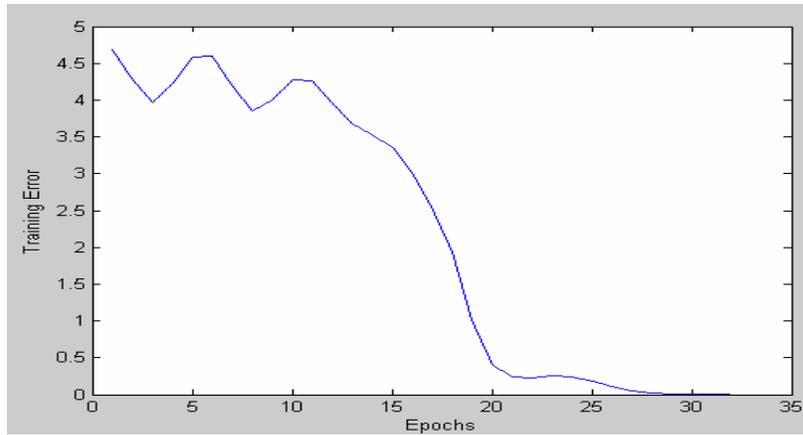

**Figure 5.2**: Learning curve of a 2-2-1 Neural Network in application to the XOR problem

The decision surface generated by the neural network is presented below:

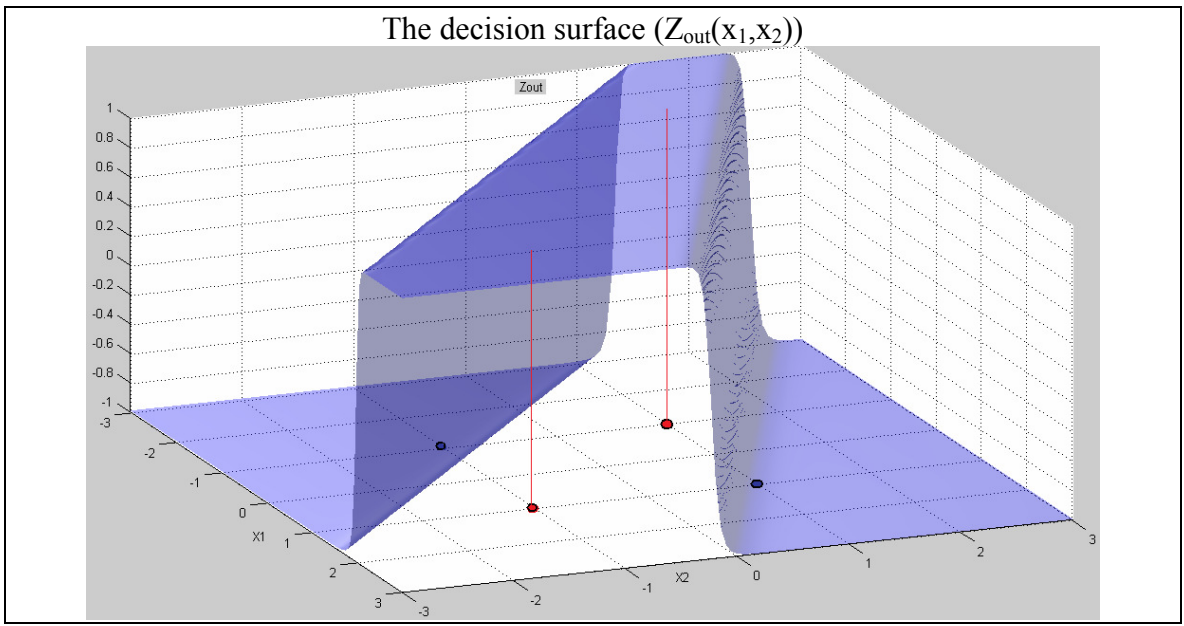

**Figure 5.2b**: Decision surface of a 2-2-1 Neural Network in application to the XOR problem

Like wise, the diagram below shows the contour diagram of the decision surface above. We can see the decision boundary is seen when the contour height is zero (straight green lines)

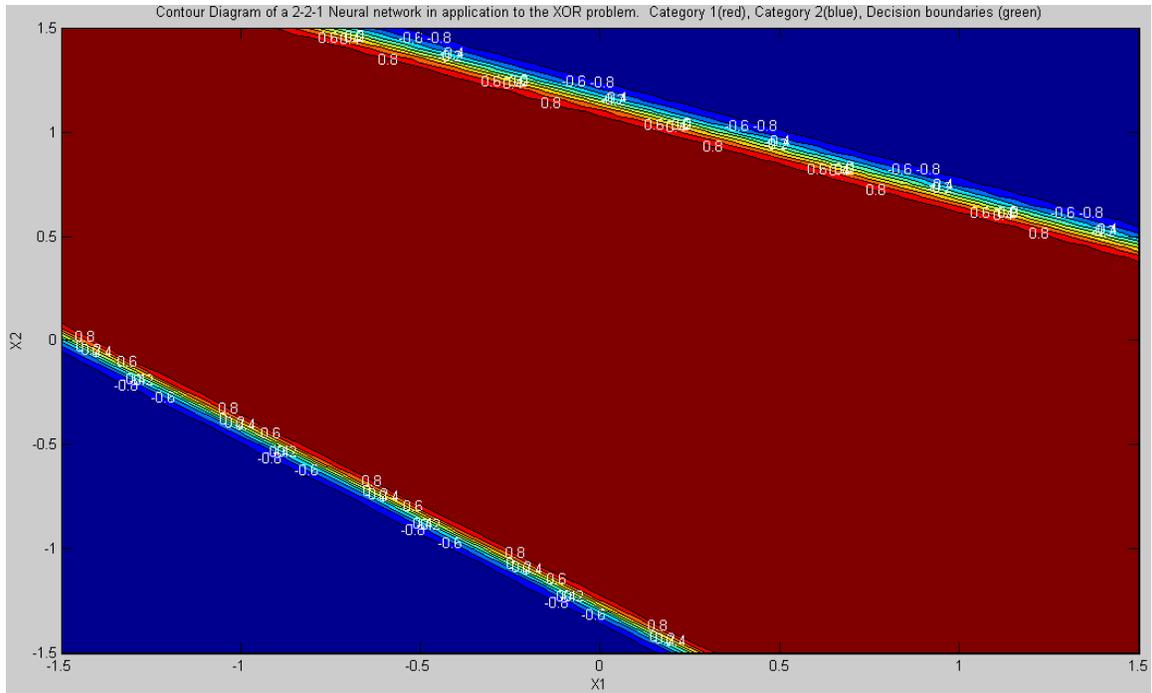
**Figure 5.2c:** Contour diagram of the 2-2-1 Neural Network in application to the XOR problem

## 5.3: Overall Comparison

Tabular comparison between the two algorithms is summarized in the table below.

**TABLE 5.3**: Classification comparison between R.R.E algorithm versus a 2-2-1 Neural Network in application to the XOR problem

| Training Points: (X1,X2) | target | R.R.E Algorithm's Z(out) (1 iteration) | Neural Network's Z(out) (32 Epochs) |
|---|---|---|---|
| $(-1,-1)^t$ | -1 | -0.9993 | -0.9887 |
| $(-1,1)^t$ | 1 | 0.9993 | 1.0 |
| $(1,-1)^t$ | 1 | 0.9993 | 1.0 |
| $(1,1)^t$ | -1 | -0.9993 | -0.9937 |

*Achieving target values*
Note it is pointless to calculate percentage error of each algorithm because the $Z_{out}$s in the table above can be made arbitrarily close to the target values by adjusting certain parameters. For the Neural Network, we can make our threshold/criterion very small and adjust our learning rate accordingly. For the R.R.E algorithm, we can use redundant training (duplication), we can also use a higher auxiliary sensitivity factor ($\lambda$=3, to achieve all $|Z_{out}|$ = 1) or alternatively, we can use a higher order variance reduction function f(n).

*Minimal training abilities*
A meaningful point to consider is that both algorithms obtain acceptable results through different mechanisms, but the R.R.E consistently requires minimal training of only one step while the Neural Network would normally require more iterations (epochs) - not unless the learning rate is perfectly on point (very rare), in which case it will require one iteration as well.

*Decision making*
In terms of decision making for a single test point, the forward feedback of a Neural Network has a computational complexity correlated with the number of weights, while the R.R.E has a computational complexity correlated with the number of training points. Usually the number of training points are 10-20 times the number of weights (adaptive parameters) in order to minimize over-fitting in a Neural Network. Therefore if we adhere to this rule, the R.R.E is generally more computationally expensive in classifying than a Neural Network of comparable complexity.

*General training*
R.R.E is non-iterative, requires minimal training and is generally computationally superior to a Neural Network during training. We have ignored the data normalization requirement of a Neural Network which provides an additional argument in favour of R.R.E.

**6 Conclusion**

We see that R.R.E is a powerful algorithm, but unlike much other learning algorithms, it is non-iterative and defers all its training computation to its discriminant function where it becomes computationally expensive while classifying.
Although R.R.E achieves stellar results on different fronts, work has to be done to reduce the computational complexity of its Discriminant function. One suggested way to resolve this is using 'training data filtering' as suggested in section 2.9.1. Another way would be to use a subset of training points/terms to be included in the Discriminant function - an intuitive suggestion would be to use the training points that have the lowest dissimilarity
with the test point in question as they have the largest
influence, but that would mean that we would have to come
up with a robust system, possibly a look-up table that can
efficiently, store, find and compare all training points to
a test point before inclusion to the Discriminant function.

## 7 Acknowledgments

I graciously thank Prof. Anastasios N. Venetsanopoulos[1]
for proofreading a preliminary version of this manuscript
in addition to offering a valuable graduate level course
(EE8209) in Intelligent Systems that equipped me with the
background to tackle this project. Three data sets used in
this project were courteously provided in the EE8209 course.

I thank Dr. Alp Kucukelbir for generously taking the time
to read this manuscript (in Sept 2014) and for suggesting
constructive comments that are yet to be implemented.

---

[1] SUBMITTED TO EE8209 COURSE INSTRUCTOR ON APRIL 27 2012

## Appendices
**Appendix B1:** "iris_setosa_versicolor" Normalized & Augmented Row Vectors

```
     1.0000     5.1000     3.5000
     1.0000     4.9000     3.0000
     1.0000     4.7000     3.2000
     1.0000     4.6000     3.1000
     1.0000     5.0000     3.6000
     1.0000     5.4000     3.9000
     1.0000     4.6000     3.4000
     1.0000     5.0000     3.4000
     1.0000     4.4000     2.9000
     1.0000     4.9000     3.1000
     1.0000     5.4000     3.7000
     1.0000     4.8000     3.4000
     1.0000     4.8000     3.0000
     1.0000     4.3000     3.0000
     1.0000     5.8000     4.0000
     1.0000     5.7000     4.4000
     1.0000     5.4000     3.9000
     1.0000     5.1000     3.5000
     1.0000     5.7000     3.8000
     1.0000     5.1000     3.8000
     1.0000     5.4000     3.4000
     1.0000     5.1000     3.7000
     1.0000     4.6000     3.6000
     1.0000     5.1000     3.3000
     1.0000     4.8000     3.4000
     1.0000     5.0000     3.0000
     1.0000     5.0000     3.4000
     1.0000     5.2000     3.5000
     1.0000     5.2000     3.4000
     1.0000     4.7000     3.2000
     1.0000     4.8000     3.1000
     1.0000     5.4000     3.4000
     1.0000     5.2000     4.1000
     1.0000     5.5000     4.2000
     1.0000     4.9000     3.1000
     1.0000     5.0000     3.2000
     1.0000     5.5000     3.5000
     1.0000     4.9000     3.1000
     1.0000     4.4000     3.0000
     1.0000     5.1000     3.4000
     1.0000     5.0000     3.5000
     1.0000     4.5000     2.3000
     1.0000     4.4000     3.2000
     1.0000     5.0000     3.5000
     1.0000     5.1000     3.8000
     1.0000     4.8000     3.0000
     1.0000     5.1000     3.8000
     1.0000     4.6000     3.2000
     1.0000     5.3000     3.7000
     1.0000     5.0000     3.3000
    -1.0000    -7.0000    -3.2000
    -1.0000    -6.4000    -3.2000
    -1.0000    -6.9000    -3.1000
    -1.0000    -5.5000    -2.3000
    -1.0000    -6.5000    -2.8000
    -1.0000    -5.7000    -2.8000
    -1.0000    -6.3000    -3.3000
    -1.0000    -4.9000    -2.4000
    -1.0000    -6.6000    -2.9000
    -1.0000    -5.2000    -2.7000
    -1.0000    -5.0000    -2.0000
    -1.0000    -5.9000    -3.0000
    -1.0000    -6.0000    -2.2000
    -1.0000    -6.1000    -2.9000
    -1.0000    -5.6000    -2.9000
    -1.0000    -6.7000    -3.1000
    -1.0000    -5.6000    -3.0000
    -1.0000    -5.8000    -2.7000
    -1.0000    -6.2000    -2.2000
    -1.0000    -5.6000    -2.5000
    -1.0000    -5.9000    -3.2000
    -1.0000    -6.1000    -2.8000
    -1.0000    -6.3000    -2.5000
    -1.0000    -6.1000    -2.8000
    -1.0000    -6.4000    -2.9000
    -1.0000    -6.6000    -3.0000
    -1.0000    -6.8000    -2.8000
    -1.0000    -6.7000    -3.0000
    -1.0000    -6.0000    -2.9000
    -1.0000    -5.7000    -2.6000
    -1.0000    -5.5000    -2.4000
    -1.0000    -5.5000    -2.4000
    -1.0000    -5.8000    -2.7000
    -1.0000    -6.0000    -2.7000
    -1.0000    -5.4000    -3.0000
    -1.0000    -6.0000    -3.4000
    -1.0000    -6.7000    -3.1000
    -1.0000    -6.3000    -2.3000
    -1.0000    -5.6000    -3.0000
    -1.0000    -5.5000    -2.5000
    -1.0000    -5.5000    -2.6000
    -1.0000    -6.1000    -3.0000
    -1.0000    -5.8000    -2.6000
    -1.0000    -5.0000    -2.3000
    -1.0000    -5.6000    -2.7000
    -1.0000    -5.7000    -3.0000
    -1.0000    -5.7000    -2.9000
    -1.0000    -6.2000    -2.9000
    -1.0000    -5.1000    -2.5000
    -1.0000    -5.7000    -2.8000        NON-NORMALIZED & UNAUGMENTED DATA SAVED AS: "iris_setosa_versicolor.mat"
```

**Appendix B2:** "iris_versicolor_virginicaV2" Normalized & Augmented Row Vectors

```
  1.0000    7.0000    3.2000
  1.0000    6.0000    3.1000
  1.0000    7.0000    3.1000
  1.0000    5.5000    2.3000
  1.0000    6.5000    2.8000
  1.0000    5.7000    2.8000
  1.0000    6.1000    3.3000
  1.0000    4.9000    2.4000
  1.0000    6.6000    2.9000
  1.0000    5.2000    2.7000
  1.0000    5.0000    2.0000
  1.0000    5.4000    3.2000
  1.0000    6.0000    2.2000
  1.0000    6.1000    2.9000
  1.0000    5.6000    2.9000
  1.0000    6.6000    3.1000
  1.0000    5.6000    3.0000
  1.0000    5.4000    2.7000
  1.0000    6.2000    2.2000
  1.0000    5.6000    2.5000
  1.0000    5.9000    3.2000
  1.0000    6.1000    2.8000
  1.0000    6.1000    2.3000
  1.0000    6.1000    2.8000
  1.0000    6.4000    2.9000
  1.0000    6.6000    3.0000
  1.0000    6.8000    2.8000
  1.0000    6.6000    2.8000
  1.0000    6.0000    2.9000
  1.0000    5.7000    2.6000
  1.0000    5.5000    2.4000
  1.0000    5.5000    2.4000
  1.0000    5.4000    2.7000
  1.0000    6.0000    2.7000
  1.0000    5.4000    3.0000
  1.0000    6.0000    3.4000
  1.0000    6.7000    2.9000
  1.0000    6.3000    2.3000
  1.0000    5.6000    3.0000
  1.0000    5.5000    2.5000
  1.0000    5.5000    2.6000
  1.0000    6.1000    3.2000
  1.0000    5.8000    2.6000
  1.0000    5.0000    2.3000
  1.0000    5.6000    2.7000
  1.0000    5.7000    3.0000
  1.0000    5.7000    2.9000
  1.0000    6.2000    2.9000
  1.0000    5.1000    2.5000
  1.0000    5.7000    2.8000
 -1.0000   -6.3000   -3.3000
 -1.0000   -5.9000   -2.8000
 -1.0000   -7.1000   -3.0000
 -1.0000   -6.3000   -2.9000
 -1.0000   -6.5000   -3.0000
 -1.0000   -7.6000   -3.0000
 -1.0000   -4.9000   -2.5000
 -1.0000   -7.3000   -2.9000
 -1.0000   -6.7000   -2.5000
 -1.0000   -7.2000   -3.6000
 -1.0000   -6.5000   -3.2000
 -1.0000   -6.4000   -2.7000
 -1.0000   -6.8000   -3.0000
 -1.0000   -5.7000   -2.5000
 -1.0000   -5.8000   -2.8000
 -1.0000   -6.4000   -3.2000
 -1.0000   -6.5000   -3.0000
 -1.0000   -7.7000   -3.8000
 -1.0000   -7.7000   -2.6000
 -1.0000   -6.0000   -2.6000
 -1.0000   -6.9000   -3.2000
 -1.0000   -5.6000   -2.8000
 -1.0000   -7.7000   -2.8000
 -1.0000   -6.3000   -2.7000
 -1.0000   -6.7000   -3.3000
 -1.0000   -7.2000   -3.2000
 -1.0000   -6.2000   -2.8000
 -1.0000   -6.1000   -3.0000
 -1.0000   -6.4000   -2.8000
 -1.0000   -7.2000   -3.0000
 -1.0000   -7.4000   -2.8000
 -1.0000   -7.9000   -3.8000
 -1.0000   -6.4000   -2.8000
 -1.0000   -6.3000   -2.8000
 -1.0000   -6.1000   -2.6000
 -1.0000   -7.7000   -3.0000
 -1.0000   -6.3000   -3.4000
 -1.0000   -6.4000   -3.1000
 -1.0000   -6.0000   -3.0000
 -1.0000   -6.9000   -3.1000
 -1.0000   -6.7000   -3.1000
 -1.0000   -6.9000   -3.1000
 -1.0000   -5.9000   -2.8000
 -1.0000   -6.8000   -3.2000
 -1.0000   -6.7000   -3.3000
 -1.0000   -6.7000   -3.0000
 -1.0000   -6.3000   -2.5000
 -1.0000   -6.5000   -3.0000
 -1.0000   -6.2000   -3.4000
 -1.0000   -5.9000   -3.0000     NON-NORMALIZED & UNAUGMENTED DATA SAVED AS: "iris_versicolor_virginicaV2.mat"
```

**Appendix B3:** "Support1" Normalized & Augmented Row Vectors

```
     1.0000     5.1000     3.5000
     1.0000     4.9000     3.0000
     1.0000     4.7000     3.2000
     1.0000     4.6000     3.1000
     1.0000     5.0000     3.6000
     1.0000     5.4000     3.9000
     1.0000     4.6000     3.4000
     1.0000     5.0000     3.4000
     1.0000     4.4000     2.9000
     1.0000     4.9000     3.1000
     1.0000     5.4000     3.7000
     1.0000     4.8000     3.4000
     1.0000     4.8000     3.0000
     1.0000     4.3000     3.0000
     1.0000     5.8000     4.0000
     1.0000     5.7000     4.4000
     1.0000     5.4000     3.9000
     1.0000     5.1000     3.5000
     1.0000     5.7000     3.8000
     1.0000     5.1000     3.8000
     1.0000     5.4000     3.6000
     1.0000     5.1000     3.7000
     1.0000     4.6000     3.6000
     1.0000     5.1000     3.3000
     1.0000     4.8000     3.4000
     1.0000     5.0000     3.0000
     1.0000     5.0000     3.4000
     1.0000     5.2000     3.5000
     1.0000     5.2000     3.4000
     1.0000     4.7000     3.2000
     1.0000     4.8000     3.1000
     1.0000     5.4000     3.3000
     1.0000     5.2000     4.1000
     1.0000     5.5000     4.2000
     1.0000     4.9000     3.1000
     1.0000     5.0000     3.2000
     1.0000     5.5000     3.5000
     1.0000     4.9000     3.1000
     1.0000     4.4000     3.0000
     1.0000     5.1000     3.4000
     1.0000     5.0000     3.5000
     1.0000     4.5000     3.0000
     1.0000     4.4000     3.2000
     1.0000     5.0000     3.5000
     1.0000     5.1000     3.8000
     1.0000     4.8000     3.0000
     1.0000     5.1000     3.8000
     1.0000     4.6000     3.2000
     1.0000     5.3000     3.7000
     1.0000     5.0000     3.3000
    -1.0000    -7.0000    -3.2000
    -1.0000    -6.4000    -3.2000
    -1.0000    -6.9000    -3.1000
    -1.0000    -5.5000    -2.3000
    -1.0000    -6.5000    -2.8000
    -1.0000    -5.7000    -2.8000
    -1.0000    -6.3000    -3.3000
    -1.0000    -4.9000    -2.4000
    -1.0000    -6.6000    -2.9000
    -1.0000    -5.2000    -2.7000
    -1.0000    -5.0000    -2.0000
    -1.0000    -5.9000    -3.0000
    -1.0000    -6.0000    -2.2000
    -1.0000    -6.1000    -2.9000
    -1.0000    -5.6000    -2.9000
    -1.0000    -6.7000    -3.1000
    -1.0000    -5.6000    -3.0000
    -1.0000    -5.8000    -2.7000
    -1.0000    -6.2000    -2.2000
    -1.0000    -5.6000    -2.5000
    -1.0000    -5.9000    -3.2000
    -1.0000    -6.1000    -2.8000
    -1.0000    -6.3000    -2.5000
    -1.0000    -6.1000    -2.8000
    -1.0000    -6.4000    -2.9000
    -1.0000    -6.6000    -3.0000
    -1.0000    -6.8000    -2.8000
    -1.0000    -6.7000    -3.0000
    -1.0000    -6.0000    -2.9000
    -1.0000    -5.7000    -2.6000
    -1.0000    -5.5000    -2.4000
    -1.0000    -5.5000    -2.4000
    -1.0000    -5.8000    -2.7000
    -1.0000    -6.0000    -2.7000
    -1.0000    -5.4000    -3.0000
    -1.0000    -6.0000    -3.4000
    -1.0000    -6.7000    -3.1000
    -1.0000    -6.3000    -2.3000
    -1.0000    -5.6000    -3.0000
    -1.0000    -5.5000    -2.5000
    -1.0000    -5.5000    -2.6000
    -1.0000    -6.1000    -3.0000
    -1.0000    -5.8000    -2.6000
    -1.0000    -5.0000    -2.3000
    -1.0000    -5.6000    -2.7000
    -1.0000    -5.7000    -3.0000
    -1.0000    -5.7000    -2.9000
    -1.0000    -6.2000    -2.9000
    -1.0000    -5.1000    -2.5000
    -1.0000    -5.7000    -2.8000          NON-NORMALIZED & UNAUGMENTED DATA SAVED AS: **"Support1.mat"**
```